\documentclass[journal]{IEEEtran}

\usepackage{enumitem}
\usepackage{amsmath,amssymb,amsfonts}
\usepackage{algorithmic}
\usepackage{graphicx}
\usepackage{physics}
\usepackage{textcomp}
\usepackage{xcolor}
\usepackage{multicol}
\usepackage{nidanfloat}

\usepackage[numbers,sort&compress,square]{natbib}
\usepackage{hyperref}
\usepackage{makecell}
\usepackage{nccmath}
\usepackage{mathtools}
\usepackage{subcaption}
\usepackage[utf8]{inputenc}
\usepackage[english]{babel}
\usepackage{amsmath, amsthm, amssymb}
\usepackage{here}

\newtheorem{theorem}{Theorem}
\theoremstyle{definition}
\newtheorem{definition}{Definition}
\theoremstyle{remark}
\newtheorem{remark}{Remark}
\theoremstyle{lemma}
\newtheorem{lemma}{Lemma}
\theoremstyle{proposition}

\theoremstyle{corollary}

\newcommand{\unit}[1] {~\textrm{#1}}

\def\BibTeX{{\rm B\kern-.05em{\sc i\kern-.025em b}\kern-.08em
T\kern-.1667em\lower.7ex\hbox{E}\kern-.125emX}}

\title{Two-Port Analysis of Stability and Transparency in \\Series Damped Elastic Actuation }

\author{Ugur Mengilli$^{*}$ \hspace{12mm} Umut Caliskan$^{*}$ \hspace{12mm} Zeynep Ozge Orhan \hspace{12mm}  Volkan Patoglu
\thanks{U. Mengilli, U. Caliskan, Z. O. Orhan and V. Patoglu are with Faculty of Engineering and Natural Sciences, Sabanc{\i} University, \.Istanbul, Turkey.\\ {\tt
\{ugurmengilli,caliskanumut,zeynepozge,volkan.patoglu\} @sabanciuniv.edu}}%
\thanks{$^{*}$ First two authors have equally contributed to this study.}
\thanks{Manuscript received \today.}}

\begin{document}

\maketitle

\begin{abstract}
Series Elastic Actuation (SEA) is a widely-used approach for interaction control, as it enables high fidelity and robust force control, improving the safety of physical human-robot interaction (pHRI).
Safety is an imperative design criterion for pHRI that limits the interaction performance since there exists a fundamental trade-off between stability robustness and rendering performance.
The safety of interaction necessitates the closed-loop stability of a pHRI system when coupled to a wide range of unknown operators and environments.
In this study, we provide the \emph{necessary and sufficient conditions for two-port passivity} of series damped elastic actuation under velocity-sourced impedance control within the frequency-domain passivity framework.
Based on the newly established conditions, we derive non-conservative passivity bounds for a virtual coupler and rigorously prove the necessity of a dissipative element parallel to the series elastic component and the necessity of a virtual coupler with dissipation for the absolute stability and two-port passivity of the system.
The additional dissipative elements in the physical filter and the virtual coupler enable the system to render virtual stiffness values higher than that can be rendered using a pure SEA\@.
Our results extend earlier studies on coupled stability by presenting the necessary and sufficient conditions for all passive terminations.
We validate our results through a set of physical experiments and systematic numerical simulations.
\end{abstract}

\vspace{-1mm}

\vspace{-3mm}

\section{Introduction} \label{sec:Introduction}

\IEEEPARstart{E}{nsuring} safe physical human-robot interactions~(pHRI) is fundamental for many applications, including service, surgical, assistive, and rehabilitation robotics.
The safety of interaction requires the impedance characteristics of the robot at the interaction port to be controlled precisely~\cite{Hogan1985a,Colgate1988phd}.
Many robotic systems rely on closed-loop force control to compensate for parasitic forces originating from their mechanical designs.
However, the performance of all closed-loop force controllers suffers from an inherent limitation imposed by the non-collocation of sensors and actuators that introduces an upper bound on the loop gain of the system~\cite{Chae1987,Eppinger1987}.

The stable loop gain of the system is mostly allocated for the force-sensing element when traditional force sensors with high stiffness are employed in the control loop.
This significantly limits the availability of high controller gains required to achieve fast response and good robustness properties.
\emph{Series elastic actuation}~(SEA) trades off large force-control bandwidth for force/impedance rendering fidelity by introducing a highly compliant force-sensing element into the closed-loop force control architecture~\cite{Howard90,Pratt1995}.
Lower force sensor stiffness allows higher gains to be utilized for responsive and robust force-control.
Moreover, the force-sensing element acts as a physical filter against impacts, impulsive loads, and high-frequency disturbances.
Possessing inherent compliance and masking the inertia of the actuator from the interaction port, SEA features favorable output impedance characteristics that are safe for human interaction over the entire frequency spectrum~\cite{Howard90,Pratt1995,Robinson1999,Sensinger2006b}.
In the literature, many SEA designs have been developed for a wide range of applications~\cite{Robinson1999,Sensinger2006,Veneman2006,Khatib2008,Kyoungchul2012,Sarac2014,Gillespie2014b,Erdogan2016,Otaran2016,Munawar2016,Woo2017,Caliskan2018,Senturk2018,Umut2020}.

\smallskip
\emph{Contributions:} In this study, we present two-port passivity and transparency analyses of SEA/SDEA under VSIC\@.
We rigorously prove the necessity of a dissipative element parallel to the series elastic component and the necessity of a virtual coupler with dissipation for the absolute stability and two-port passivity of the system.
The additional dissipative elements in the physical filter of SEA and the virtual coupler enable the system to render virtual stiffness values higher than that can be rendered using a pure SEA\@.
Our results extend earlier studies on coupled stability of SEA/SDEA by presenting the \emph{necessary and sufficient conditions for all passive terminations}.

We present an analytical method using Sturm's Theorem for analyzing the positive realness of impedance transfer functions. We show that a feed-forward action canceling the measured interaction force may deteriorate the coupled stability of the system. Furthermore, we prove that the integral gain of the motion controller is required to ensure passivity. We validate our theoretical results through numerical simulations and by reproducing one-port passivity results as special cases under appropriate terminations.

\smallskip
\emph{Outline:} The rest of the paper is organized as follows:
Section~\ref{sec:RelatedWork} provides a comprehensive overview of related works addressing coupled stability of SEA/SDEA. Section~\ref{sec:Preliminaries} reviews the fundamental concepts utilized in this study.
Section~\ref{sec:SystemDescription} describes the system model and presents the need for a virtual coupler to ensure coupled stability of the system while interacting with all passive environments.
Section~\ref{sec:TwoportPassivityAnalysis} provides the necessary and sufficient conditions for the two-port passivity of the system model and proves the necessity of a parallel damper in both the physical filter and the virtual coupler.
Section~\ref{sec:PerformanceAnalysisofSDEA} analyzes the performance of the system based on its two-port model.
Section~\ref{sec:Numerical-Evaluations} numerically studies the theoretical results and compares the passivity bounds with those derived from unconditional stability.
Section~\ref{sec:Discussion} provides a discussion of the results by comparing them with related work, while Section~\ref{sec:Conclusion} concludes the paper.

\section{Related Work} \label{sec:RelatedWork}

The performance of SEA depends synergistically on its mechatronic design and controller~\cite{Kamadan2017,Kamadan2018}.
The high-performance controller design for SEA to be used in pHRI is challenging since ensuring the safety of interactions is an imperative design requirement that constrains the design process.
Safety of interaction requires coupled stability of the controlled SEA together with a human operator;
however, the presence of a human operator in the control loop significantly complicates the stability analysis.
In particular, a comprehensive model for human dynamics is not available, as human dynamics is highly nonlinear and time and configuration dependent.
Contact interactions with the environment also pose similar challenges, since the impedance of the contact environment is, in general, uncertain.

The coupled stability analysis of robotic systems in the absence of human and environment models is commonly conducted using the frequency-domain passivity framework~\cite{Colgate1988}.
In this approach, even if the human operator behaves actively, coupled stability can still be concluded through the passivity analysis, as long as the human behavior is assumed to be non-malicious.
Furthermore, non-animated environments are passive.
Therefore, coupled stability of the overall system can be concluded, if the closed-loop SEA with its controller can be designed to be passive.

While the frequency-domain passivity paradigm provides robust stability for a broad range of human and environment models, results derived from such analysis may be conservative. Less conservative paradigms, such as time-domain passivity~\cite{hannafordRyu,ryuGeneral}, complementary stability~\cite{Buerger2007,Aydin2018}, bounded-impedance absolute stability~\cite{Haddadi2010,Willaert2011,Lee2019}, may be utilized to achieve better performance while still ensuring coupled stability of interaction. Although these techniques are highly valuable, they are limited in that they rely on numerical computations/optimizations; hence, they cannot provide closed-form analytical solutions and general insights. The frequency-domain passivity analyses are highly valuable as they provide a fundamental understanding of the underlying trade-offs governing the dynamics of the closed-loop system.

\vspace{-3mm}
\subsection{Coupled Stability of SEA}

Coupled stability of SEA, modeled as an LTI system, has been investigated extensively using one-port passivity analysis, under several control architectures. Among these SEA control architectures, velocity-sourced impedance control~(VSIC)~\cite{Wyeth2008} has been favored in the literature, due to its robustness, high performance, and ease of parameter tuning~\cite{Howard90,Robinson1999,Pratt2004,Wyeth2008,Umut2020,FatihEmre2020}. 

Vallery~\textit{et~al.}~\cite{Vallery2007} have analyzed the passivity of VSIC architecture, without the motor damping in SEA model, for the case of zero reference torque. They have suggested conservative \emph{sufficient} conditions for passivity based on the actuator inertia and a ratio between the controller gains. Later, they have extended this result for stiffness rendering with VSIC and proved that the passively renderable virtual stiffness is bounded by the stiffness of physical spring in the SEA~\cite{Vallery2008}.

Tagliamonte~\textit{et~al.}~\cite{Tagliamonte2014b} have shown that less conservative \emph{sufficient} conditions for passivity can be derived for null impedance and pure stiffness rendering with VSIC architecture when the motor damping is included in the SEA model. In particular, it has been proven that the maximum achievable stiffness is not only related to the physical stiffness of the SEA, but also the physical damping in the system. They have demonstrated that the Voigt model, which is a linear spring-damper pair in parallel, cannot be passively rendered using VSIC architecture. Later, they have also shown that the Maxwell model, which is a linear spring-damper pair in series, can be passively rendered using VSIC architecture~\cite{Tagliamonte2014b}, and derived \emph{sufficient} conditions to characterize the range of environment parameters that preserve passivity. 

Calanca~\textit{et~al.}~\cite{Fiorini2017} have derived \emph{sufficient} conditions for the passivity of SEA under several control architectures: basic impedance control, VSIC, collocated admittance control, and collocated impedance control. They have shown that the limitation on maximum achievable stiffness to render a pure stiffness, as derived in~\cite{Vallery2008}, also holds for these controllers. These theoretical analyses rely on the use of non-causal differentiator terms for the force controller and neglect the effect of motor damping in the system model. It is also stated in~\cite{Fiorini2017} that the Voigt model cannot be passively rendered with VSIC architecture and an impedance controller with ideal acceleration feedback has been suggested. Theoretically, ideal acceleration feedback can be used to cancel out the influence of load dynamics; however, noise and bandwidth restrictions of acceleration signals and potential overestimation of feed-forward signals resulting in feedback inversion are important practical challenges that have limited the adaptation of the acceleration-based control, since initially proposed in~\cite{Pratt1995,Robinson1999}.

Tosun and~Patoglu~\cite{FatihEmre2020} have derived the \emph{necessary and sufficient} conditions for the passivity of VSIC architecture of SEA, relaxing the earlier established sufficiency bounds and extending the range of impedances that can be passively rendered. They have shown the \emph{necessity} of integral gain of the motion controller to render pure stiffness. Furthermore, they have proven the \emph{necessity} of a bound on the integral gains due to the inevitable physical damping in the system. This counter-intuitive bound indicates that the motor damping reduces the dynamic range of passively renderable impedances.

\vspace{-3mm}
\subsection{Physically Damped SEA}

The main disadvantage of SEA is significantly decreased large force bandwidth caused by the increase of the sensor compliance under actuator saturation~\cite{Pratt1995}.
The selection of appropriate stiffness of the compliant element is essential in SEA designs, where a compromise solution needs to be reached between force control fidelity and large force bandwidth.
Possible high-frequency oscillations of the end-effector, especially when the SEA is not in contact and the potential energy storage by the elastic element may pose as other challenges of SEA designs.

To address these issues, Newman has proposed a mechanical filter in the form of a parallel spring-damper~\cite{Newman1992}.
He has also shown that the insertion of the damper can relax the passivity bounds of the system at frequencies greater than the natural frequency of the filter, and proposed a controller, called Natural Admittance Controller, guaranteeing the passivity of the system.
Later, Dohring and Newman have further investigated the improvements of this filter on the system performance, especially at high frequencies~\cite{Dohring2002}.

The use of a physical damper instead of the series elastic element has been proposed in~\cite{Chew2004} to achieve similar improvements over SEA\@.
It has been argued that \emph{series damper actuator}~(SDA) is favorable for force control, as it features an adequate level of force fidelity, low output impedance, and a large force range.
Furthermore, it is shown through a theoretical analysis that SDA may increase the control bandwidth of the system, as it possesses a lower relative order in its transfer function compared to that of SEA\@.

Physically damped SEA concept has been studied in several other works in the literature~\cite{Hurst2004,Oblak2011,Garcia2011,Laffranchi2011,Laffranchi2014,Ott2017}.
It is argued in~\cite{Hurst2004} that impact forces may cause instability and chatter in SEA since the rapid accelerations cannot be achieved due to the rotor inertia and the motor torque limits.
It has been shown through numerical simulations that \emph{series damped elastic actuator}~(SDEA) can increase the force control bandwidth.

To improve efficiency by avoiding continual energy dissipation due to constant damping, SDEA with semi-active and variable damping have also been proposed. For instance, in~\cite{Garcia2011}, SDEA has been implemented for a legged robot using a magneto-rheological brake, where SDEA is controlled with a cascaded control architecture that has an inner force control loop and outer position control loop. Through physical experiments, it has been shown that adding parallel damping reduces oscillations and improves energy consumption.

In~\cite{Laffranchi2011}, the importance of damping to reduce oscillations has been highlighted, in the context of variable stiffness actuation. Within an admittance control scheme, it is stated that the introduction of damping acts as a phase lead after the resonance, resulting in improvements in the stability of the system. In~\cite{Laffranchi2014}, it has been shown that admittance controlled SDEA can achieve the same dynamic control performance of a conventional SEA, but with less effort, particularly for systems with a low natural frequency. It is also stated that although stability and control performance are enhanced, the level of actuator safety is compromised due to the increase of the transmitted force with the addition of damping.

In~\cite{Focchi2016}, numerical stability maps have been used to determine the viable range of stiffness
and damping values for SDEA under a cascaded impedance controller with an inner torque loop acting on a
velocity-compensated plant and load dynamics. Velocity compensation is implemented using a positive velocity feedback loop that aims to increase the bandwidth of the torque loop under passivity constraints.

In~\cite{Ott2017}, conventional SEA and physical damped SEA structures have been compared from a control design perspective. The role of natural velocity feedback effect on force control performance is discussed, and the addition of physical damping to reduce the relative order of force dynamics is advocated. It is shown that through the addition of damping into SEA structure, derivative (D-control) terms become no longer necessary for the force control; therefore, acceleration feedback can be avoided. This study also suggests that robustness of SEA against impacts can be recovered by SDEA, if the transmitted damper force is mechanically limited, for instance, through a slip clutch.

While these studies present advantages of SDEA over SEA, in terms of energy efficiency, reduction of the oscillations and lack of need for D-control terms, they have not addressed the coupled stability of interaction with SDEA during impedance rendering. In~\cite{Oblak2011}, one-port passivity analysis for SDEA under basic impedance control has been presented. The control architecture utilized in this study is somewhat unconventional; in addition to the series damped elastic element, another force sensor is utilized after the end-effector inertia for measuring the human force, and this interaction force is fed back to the controller. In this study, Oblak and Matjačić~\cite{Oblak2011} have shown that adequate level of mechanical damping in the compliant element is needed to ensure the passivity of pure stiffness rendering. Moreover, \emph{sufficient} conditions for passivity and the lower bound on the required physical damping have been derived, in terms of the controller gains, motor-side damping, end-effector inertia, and motor-side inertia. While passivity of SEA/SDEA is independent of the end-effector inertia under conventional controllers, in this work, the use of the second force sensor after the end-effector introduces an additional bound on the proportional force controller gain that depends on the ratio of the actuator to the end-effector inertia.

\vspace{-3mm}
\subsection{Two-Port Analyses of Physically Damped SEA}

In the literature, several works have conducted the coupled stability analysis of SEA/SDEA using a one-port passivity analysis, where the environments have a certain form.
On the other hand, Tognetti~\cite{Tognetti2005} has studied a general haptic device with various virtual coupler forms considering two-port absolute stability.
He has attempted to improve stability by inserting additional damping on the motor side.
However, his analysis is numerical and does not provide general insights.
In this study, we propose an analytical two-port passivity analysis of SDEA under VSIC with a virtual coupler.
Two-port modeling provides an analysis framework that is advantageous in several ways:

\begin{itemize}
	\item[i)] Two-port passivity analysis ensures that the controller of SEA/SDEA can be designed independent from the environment to be rendered, as the coupled stability of the system can be ensured for any passive terminating environment. This may be especially useful if the environment characteristics are unknown to the device/controller designer, as commonly the case in haptic rendering~\cite{Adams99}, where the dynamics of the virtual environment rely on some external simulator.

	\item[ii)] Using two-port analysis provides direct correspondence with the bilateral teleoperation literature and enables the passivity/transparency of the system to be analyzed analogously. For instance, the two-port representation of a haptic device provides an elegant way to observe the velocity/force transmission between the operator and the (virtual) environment.

	\item[iii)]  Two-port analyses provide more general solutions, from which one-port results may be derived by properly terminating two-port element with an appropriate passive one-port~\cite{Haykin70}.
\end{itemize}

\vspace{-3mm}
\section{Preliminaries} \label{sec:Preliminaries}

\subsection{Two-Port Network Representation}

Two-port models have been adopted by the robotics to analyze the coupled stability~\cite{Anderson89} and performance~\cite{Hannaford1989} of haptic and bilateral teleoperation systems through the energy exchange analogy to the circuit theory. A two-port element can be represented by an \emph{immittance matrix} that captures the relation between the effort $(F_1, F_2)$ and the flow variables $(v_1, v_2)$.
It is common practice for efforts to represent voltages or forces, while the corresponding flows are considered as currents or velocities.

Six distinct immittance matrices can be expressed based on the selection of the independent variables of the ports~\cite{Haykin70}.
It is favorable to use the hybrid matrix (or \emph{h}-matrix) when the input velocity and the output force are available~\cite{Hannaford1989}.
By selecting these two as the independent variables, the following matrix relation describes the two-port element.

\begin{equation}
	\begin{bmatrix}
		F_1 \\
		v_{2}
	\end{bmatrix}
	=
	\begin{bmatrix}
		h_{11} & h_{12} \\
		h_{21} & h_{22}
	\end{bmatrix}
	\begin{bmatrix}
		v_1 \\
		F_2
	\end{bmatrix}.
\end{equation}
In this form, $h_{11}$ and $h_{22}$ terms reveal important stability characteristics of the input and output ports.
On the other hand, $h_{12}$ ({reverse force-transfer ratio}) and $h_{21}$ ({forward velocity-transfer ratio}) provide insights about the performance of the network.

\vspace{-3mm}
\subsection{Coupled Stability}

The coupling of two stable systems does not necessarily result in a stable overall system since the dynamics of interaction is also important for stability.
pHRI often demands robust stability while interacting with a wide range of environment and operator dynamics whose models are not available.
Colgate and Hogan~\cite{Colgate1988} have proposed the \emph{frequency-domain passivity} to address the stability of interconnected systems.

Systems that do not produce energy are passive;
hence, they are inherently stable.
A useful property of passivity is that parallel and negative feedback interconnections of two passive systems also result in a passive system.
This property can be used to ensure the stability of an interconnected system.
Along these lines, \emph{coupled stability} is defined as follows:

\begin{definition}[Coupled Stability~\cite{Fasse1987, Colgate1988phd}]
	\label{def:coupled-stability}
	A system has coupled stability property if:
	\begin{enumerate}[label=\roman*)]
		\item The system is stable when isolated.
		\item The system remains stable when coupled to any passive environment that is also stable when isolated.
	\end{enumerate}
\end{definition}

In frequency-domain passivity analysis, in general, the human operator is not assumed to be passive but is required to be non-malicious, i.e., does not aim to destabilize the system deliberately.
For such interaction, human-applied inputs can be modeled to have a passive component and an intentionally applied active component that can be assumed to be independent of the system states.
Given that state-independent active terms do not violate the coupled stability conclusions of the frequency-domain passivity, coupled stability can be concluded as if the human operator is passive when the state-independent active terms are neglected~\cite{Colgate1988}.

\smallskip
\subsubsection{One-Port Passivity}

Given a one-port, LTI, stable plant coupled to a passive environment, a necessary and sufficient condition for the coupled stability (see Definition~\ref{def:coupled-stability}) of the system is that the one-port is passive~\cite{Colgate1988phd}.
The driving point impedance $Z(s)$ of a one-port LTI network is passive if and only if it is positive real~\cite{Haykin70, Colgate1988phd}.

\begin{theorem}[Positive Realness~\cite{Haykin70}]
	\label{thm:1port-pass}
	An impedance function $Z(s)$ is positive real \textbf{if and only if}:
	\begin{enumerate}
		\item $Z(s)$ has no poles in the right half plane.
		\item Any poles of $Z(s)$ on the imaginary axis are simple with positive and real residues.
		\item $\emph{Re}\left[Z(j\omega)\right] \geq 0$ for all $\omega$.
	\end{enumerate}
\end{theorem}

\begin{figure}[t!]
	\centering
	{\includegraphics[width=.9\columnwidth]{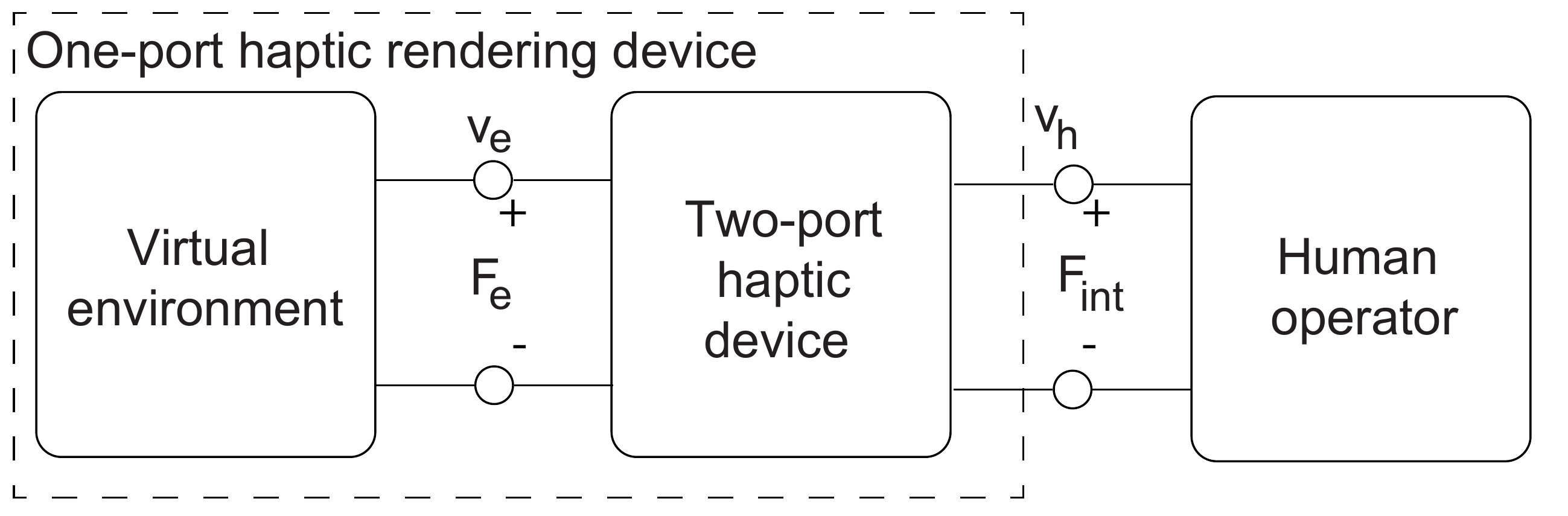}}
	\vspace{-.5\baselineskip}
	\caption{Network representations of a haptic rendering}
	\vspace{-.85\baselineskip}
	\label{fig:haptic-network}
\end{figure}

During haptic rendering, a human operator interacts with a virtual environment (VE) through the controlled device, as shown in Figure~\ref{fig:haptic-network}.
In this figure, the operator and the VE can be considered as one-port elements, while the controlled device (excluding the VE dynamics) can be considered as a two-port network.

When the model of the VE to be rendered is known, then the two-port controlled device model can be terminated with this specific environment to form the one-port rendering model, as depicted by the dashed lines in Figure~\ref{fig:haptic-network}.
For this one-port model, the driving point impedance of the interaction port can be defined as:
\begin{equation}
	\label{eq:def-impedance}
	Z(s) = \frac{F_{\text{int}}}{v_h},
\end{equation}
\noindent where $F_{\text{int}}$ represents the interaction force between the operator and the end-effector of the device and $v_h$ denotes their mutual velocity.
Ensuring passivity of the driving point impedance $Z(s)$ ensures coupled stability of interactions with a non-malicious human operator.

\smallskip
\subsubsection{Two-Port Passivity}

Two-port passivity analysis considers the controlled device, excluding the VE and human operator dynamics.
Both the VE and human operator are assumed to be passive (with any active components being state independent).
A sufficient condition of coupled stability of the overall system is the passivity of the two-port element.
Note that two-port passivity is a conservative means of ensuring coupled stability.
The necessary and sufficient conditions for the passivity of an LTI two-port element characterized by an immittance matrix $H$ are given as follows:

\begin{theorem}[Two-Port Passivity~\cite{Haykin70}]
	\label{thm:2port-pass}
	A linear time-invariant (LTI) two-port network is passive \textbf{if and only if}:
	\begin{enumerate}
		[label=(\alph*)]
		\item The \emph{h}-parameters have no poles in the right half plane.
		\item Any poles of the \emph{h}-parameters on the imaginary axis are simple, and the residues are real and positive.
		\item The \emph{h}-parameters satisfy the following conditions for all~$\omega$.
		\begin{enumerate}
			[label=(\roman*)]
			\item $\emph{Re}(h_{11}) \geq 0$ and $\emph{Re}(h_{22}) \geq 0$,
			\item $\emph{Re}(h_{11}) \emph{Re}(h_{22})-\abs{\frac{h_{12}^* + h_{21}}{2}}^2\geq0$.
		\end{enumerate}
	\end{enumerate}
\end{theorem}

\subsubsection{Absolute Stability}

When a two-port network remains stable under all possible passive terminations, it is said to be absolutely or unconditionally stable~\cite{Haykin70}.
Absolute stability is less conservative condition compared to two-port passivity.
The necessary and sufficient conditions for absolute stability of an LTI two-port element characterized with an immittance matrix $H$ can be expressed as follows:

\begin{theorem}[Llewellyn's Absolute Stability~\cite{Haykin70}]
	\label{thm:abs-stability}
	A linear time-invariant (LTI) two-port network is absolutely stable \textbf{if and only if}:
	\begin{enumerate}
		[label=(\alph*)]
		\item The \emph{h}-parameters have no poles in the right half plane.
		\item Any poles of the \emph{h}-parameters on the imaginary axis are simple, and the residues are real and positive.
		\item \emph{h}-parameters satisfy the following conditions for all~$\omega$.
		\begin{enumerate}
			[label=(\roman*)]
			\item $\emph{Re}(h_{11}) \geq 0$,
			\item $2\emph{Re}(h_{11}) \emph{Re}(h_{22})-\emph{Re}(h_{12} h_{21}) - \abs{h_{12} h_{21}} \geq 0$.
		\end{enumerate}
	\end{enumerate}
\end{theorem}

\vspace{-3mm}
\subsection{Sturm's Theorem for Positiveness of a Polynomial}
Positive realness of an impedance function, as required in Condition \emph{(c)} of Theorems~\ref{thm:2port-pass} and~\ref{thm:abs-stability}, is commonly reduced to an equivalent problem of the positiveness of a polynomial by invoking the following Lemma.
\begin{lemma}
	\label{lem:ReHs}
	Let $H(s)$ be any real-rational function such that
	\begin{equation}
		\vspace{-2mm}
		\label{eq:propHs}
		H(s) = N(s) / D(s).
	\end{equation} \vspace{-1mm}
	Positive realness of $ H(s) $ can be inferred from the positiveness of the following real polynomial. \vspace{-2mm}
	\begin{IEEEeqnarray}{rCl}
		\mathrm{Re}(N(j\omega) D(-j\omega)) &=& \sum_{i=0}^{n} c_{i} \omega^{i} \geq 0, \,\,\, \forall \omega, c_{i} \in \real. \nonumber
	\end{IEEEeqnarray}
\end{lemma} \vspace{-1mm}
\begin{IEEEproof}
	The proof is trivial and has been presented in several earlier works, including~\cite{FatihEmre2020}.
\end{IEEEproof} \vspace{1mm}
Analytical solutions to establish positiveness of a polynomial are well-established for polynomials of up to degree three~\cite{Chen2009}.
However, establishing such analytical solutions becomes difficult for higher-order polynomials.
Sturm, and later Vincent, have proposed simplified solutions to this problem by decomposing polynomials of degree $n$ into the evaluation of lower degree sequences~\cite{Akritas2010}.
While Vincent's theorem is more efficient for numerical evaluations, we favor Sturm's theorem as it can also provide \emph{analytical} bounds of positiveness of a polynomial.

\begin{definition}[Sturm's Sequence or Chain~\cite{Akritas2010}]
	\label{def:sturm-seq}
	Let $f(x) = 0$ be a polynomial equation of degree $n$, with rational coefficients and without multiple roots.
	The \emph{Sturm sequence} is
	\begin{equation}
		\label{eq:sturm-seq}
		S_{seq} (x) = \{f(x), f'(x), r_1(x), r_2(x),\ldots, r_k(x)\},
	\end{equation}
	\noindent where $f'(x)$ is the first derivative of $f(x)$ and the polynomials $r_i (x), 1 \leq i \leq k \leq n-1$, are the \emph{negatives} of the remainders obtained by applying the Euclidean greater common divisor algorithm on $f(x)$ and $f'(x)$, such that
	%
	\begin{IEEEeqnarray}{rcl}
		f(x) &=& f'(x) q_1(x) - r_1(x) \nonumber\\
		f'(x) &=& r_1(x) q_2(x) - r_2(x) \nonumber\\
		&\vdots& \nonumber\\
		r_{k-2} &=& r_{k-1}(x) q_k(x) - r_k(x). \nonumber
	\end{IEEEeqnarray}
\end{definition}

\begin{theorem}[Sturm's Theorem of 1829 for real roots~\cite{Akritas2010}]
	Let $f(x) = 0$ be a polynomial equation of degree $n$, with rational coefficients and without multiple roots.
	Then, the number $\rho$ of its real roots in the open interval $(a, b)$ satisfies the equality $\rho = \nu_a - \nu_b$, where $\nu_a$, $\nu_b$ are the number of sign variations in the Sturm sequence $S_{\textrm{seq}(a)},\,S_{\textrm{seq}(b)}$, respectively.
\end{theorem}

Utilizing Sturm's theorem, the positiveness of the polynomial $f(x)$ can be found by setting the $a$ and $b$ to cover the entire frequency range, i.e., $\omega \in (-\infty, \infty)$.

\vspace{-3mm}
\subsection{Performance} \label{subsec:prelim-performance}

In addition to analysing coupled stability, two-port representation is also useful to study the performance of pHRI systems.
A commonly used concept in haptic rendering and bilateral teleoperation literature is \emph{transparency}, which quantifies the match between the mechanical impedance of the VE and that felt by the human operator, with the requirement of identical force/velocity responses.
For a two-port system represented by its hybrid immittance matrix, ideal transparency is defined as~\cite{Hannaford1989, Hashtrudi-Zaad2001}:
\begin{equation}
	\label{eq:trans}
	\begin{bmatrix}
		F_1 \\ -v_{2}
	\end{bmatrix}
	=
	\begin{bmatrix}
		0 & 1 \\
		-1 &  0
	\end{bmatrix}
	\begin{bmatrix}
		v_1 \\ F_2
	\end{bmatrix}.
\end{equation}

If $Z_{e}$ characterize the impedance of the VE, the impedance transmitted to the operator $Z_{\text{to}}$ can be computed in terms of the parameters of the hybrid matrix as~\cite{Haykin70, Hashtrudi-Zaad2001}
\begin{equation}
	\label{eq:zto}
	Z_{\text{to}} = \frac{h_{11} + \Delta_h Z_e}{1 + h_{22} Z_e},
\end{equation}
\noindent where $\Delta_h = h_{11} h_{22} - h_{12} h_{21}$.

The difference between the minimum and the maximum achievable impedances of $Z_{\text{to}}$ defines the range of passively renderable impedances, called $Z_{\text{width}}$~\cite{Colgatezwidth94}.
In terms of hybrid matrix parameters, $Z_{\min}$ and $Z_{\text{width}}$ can be computed as:
\begin{IEEEeqnarray}{rCl}
	&Z_{\min} &= h_{11}\IEEEyesnumber\IEEEnosubnumber\\
	&Z_{\text{width}} &= -\frac{h_{12} h_{21}}{h_{22}}. \IEEEyesnumber\IEEEnosubnumber
\end{IEEEeqnarray}
\vspace{-2mm}

\vspace{-3mm}

\section{System Description} \label{sec:SystemDescription}

In this section, we present the dynamic modeling of SDEA under velocity-sourced impedance control~(VSIC).

\vspace{-3mm}

\subsection{Uncontrolled SDEA Plant}

The dynamic model of the SDEA consists of the actuator mass~$M$ and associated viscous friction~$B$, which includes the effects of transmission and electrical damping due to the resistance in the actuator.
A spring~$K_f$ that obeys the linear Hooke's law, and a viscous damper~$B_f$ connect the actuator and the end-effector.
The actuator and end-effector velocities are denoted by $v$ and $v_h$, respectively.
The interaction force~$F_{\text{int}}$ is the sum of the forces induced on the linear spring and the viscous damper, which act in parallel. 

\begin{figure*}[t]
	\centering
	{\includegraphics[width=\textwidth]{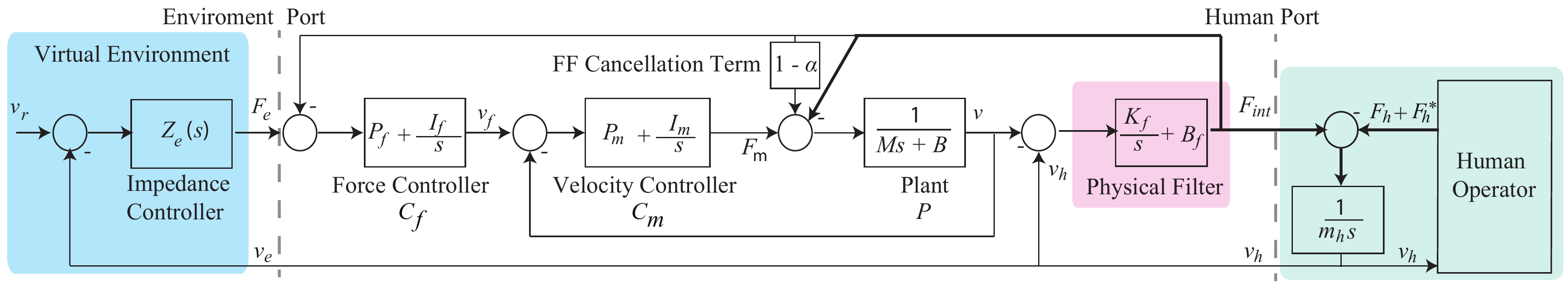}}
	\vspace{-1.5\baselineskip}
	\caption{Block diagram of SDEA under velocity-sourced impedance control~(VSIC) coupled to a human operator} \label{fig:controldiagram}
	\vspace{-.75\baselineskip}
\end{figure*}

\vspace{-2mm}

\subsection{Velocity-Sourced Impedance Control of SDEA}

Figure~\ref{fig:controldiagram} depicts the block diagram of SDEA under VSIC, where the physical interaction forces are denoted by thick lines.
In particular, the cascaded controller comprises an inner velocity and an outer force control loops.
While the inner loop renders the system into an \emph{ideal motion source}, the outer loop generates references for the velocity controller such that the spring-damper deflections are at the desired level to match the reference force.
To counteract steady-state errors, both velocity and force control loops employ PI controllers with gains denoted by $P_m$--$I_m$ and $P_f$--$I_f$, respectively.
The controllers do not include derivative action.
Given that ideal differentiation is non-causal, filters that regulate the high-frequency phase response of the controller need to be considered for the soundness of the theoretical analysis.
Besides, noise in force signals is known to significantly limit the practical use of derivative terms.
Optionally, a feed-forward signal appends to the control signal to compensate for a portion (set by $1 - \alpha$ for $0 \leq \alpha \leq 1$) of the interaction force.
The outermost loop implements an impedance controller to generate references to the force controller to display the desired impedance~$Z_e$ around the equilibrium~$v_r$ of the VE.
The effects of these parameters on the desired impedance rendering are discussed in detail in Sections~\ref{subsec:numerical-performance} and~\ref{subsec:discussion-Effectofvirtualcoupler}.
The forces applied by the human operator are modelled to have two distinct components: $F_h$ representing the passive component and $F_h^*$ denoting the intentionally applied active component that is assumed to be independent of the system states.
The end-effector mass is denoted by~$m_h$.

\vspace{-3mm}

\subsection{Simplifying Assumptions}

Following simplifying assumptions are considered:

\begin{itemize}
	\item Nonlinear effects, such as stiction, backlash, and motor saturation are neglected to develop a linear time-invariant (LTI) model. In the literature, it has been demonstrated that the cascaded force-velocity control scheme can effectively compensate for stiction and backlash~\cite{Sensinger2006,Wyeth2008}. If the motor is operated within its linear range, then the other nonlinear effects, like motor saturation, also vanish.
	\item The electrical dynamics of the system is approximated based on the commonly employed assumption that electrical time constant of the system is orders of magnitude faster than the mechanical time constant.
	\item The motor velocity signal and the rate of change of deflections on the physical filter are available with a negligible delay. For motors furnished with high-resolution encoders, differentiation filters running at high sampling frequencies (commonly on hardware) can be employed to result in velocity estimations with minimum delays, within the bandwidth of interest.
	\item Human interactions are non-malicious and do not aim to destabilize the system deliberately. In particular, human applied inputs are modelled to have a passive component and an intentionally applied active component that is assumed to be independent of the system states~\cite{Colgate1988}. This is a commonly employed assumption in the frequency-domain passivity analysis.
	\item For simplicity of analysis and without loss of generality, the VE is assumed to be grounded.
\end{itemize}

\vspace{-3mm}
\subsection{Two-Port Model of VSIC of SDEA}

To analyze the coupled stability of SDEA under VSIC, we model the closed-loop system as a two-port element that is terminated by a human operator at one-port and a passive (virtual) environment at the other port.

Selecting the input/output relationship to correspond to that of a hybrid immittance matrix, the two-port model can be expressed as:
\begin{equation}
	\label{eq:Hmatrix}
	\begin{bmatrix}
		F_{\text{int}} \\ v_e
	\end{bmatrix}
	=
	\begin{bmatrix}
		h_{11} & h_{12}    \\
		-1 & 0
	\end{bmatrix}
	\begin{bmatrix}
		-v_h \\    F_e
	\end{bmatrix},
\end{equation}
\noindent where \vspace{-2mm}
\begin{equation*}
	h_{11} = {\frac{(K_f + B_f s)(C_m P + 1)}{(C_m P + 1)s + P(K_f + B_f s)(C_m C_f + 1)}}
\end{equation*}
\begin{equation*}
	h_{12} = {\frac{C_f C_m P(K_f + B_f s)}{(C_m P + 1) s + P (K_f + B_f s)(C_m C_f + 1)}}.
\end{equation*}

In this representation, $P$ denotes the actuator dynamics, $C_m$ and $C_f$ denote generic motion and force controllers, respectively.

\begin{figure}[h]
	\centering
	{\includegraphics[width=.9\columnwidth]{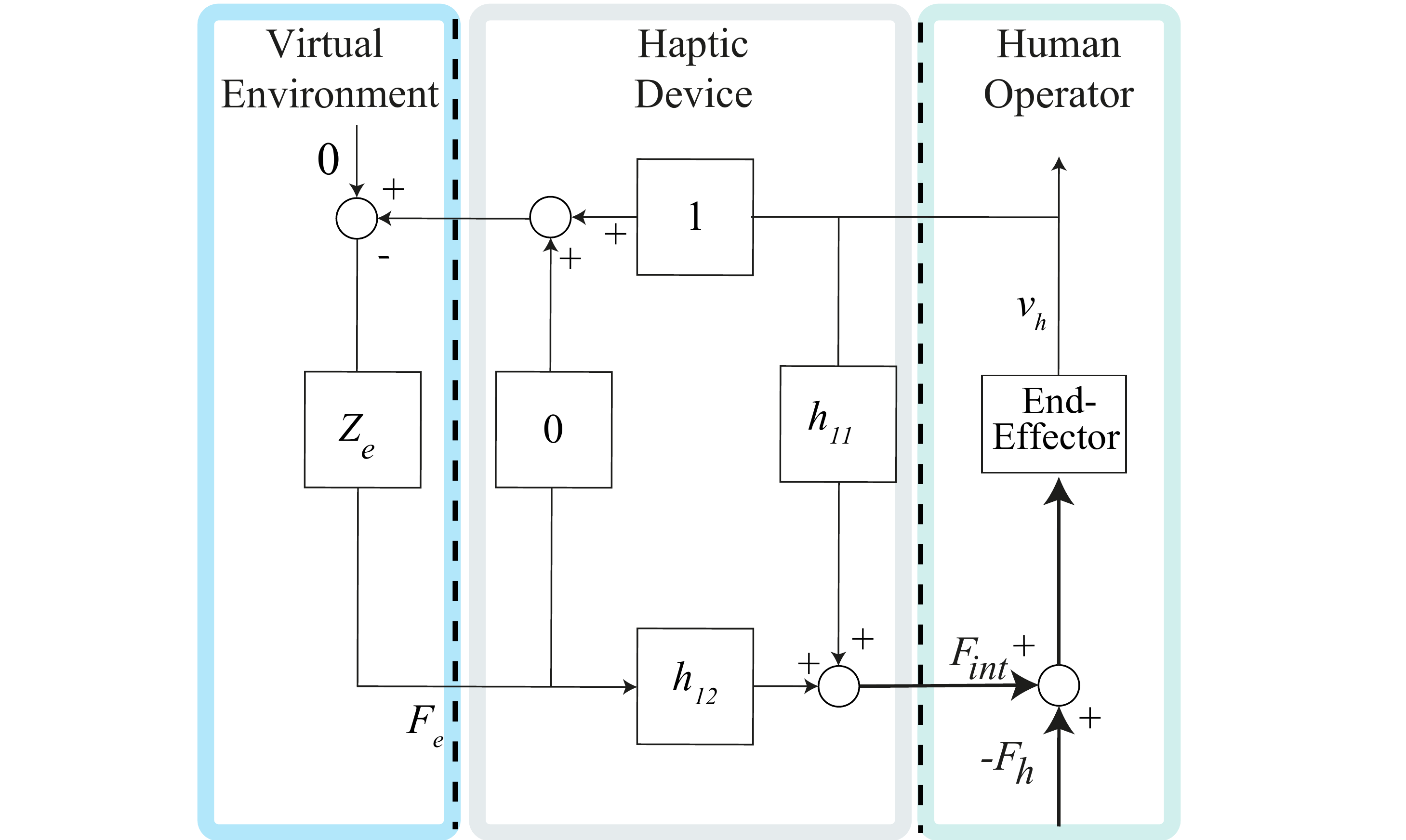}}
	\vspace{-.5\baselineskip}
	\caption{Re-arranged block diagram of SDEA under VSIC that explicitly depicts the underlying two-port model}
	\label{fig:generalizedSEA}
	\vspace{-.8\baselineskip}
\end{figure}

Figure~\ref{fig:generalizedSEA} presents a re-arrangement of the block diagram in Figure~\ref{fig:controldiagram}, such that the underlying two-port model becomes explicit. For one to one correspondence with Figure~\ref{fig:controldiagram}, one can set $P = \frac{1}{M s + B}$, $C_m = P_m + \frac{I_m}{s}$, and $C_f = P_f + \frac{I_f}{s}$.

\vspace{-3mm}

\subsection{Coupled Stability of VSIC of SDEA}

The two-port model of controlled SDEA, as given in Eqn.~\eqref{eq:Hmatrix}, is neither two-port passive nor absolutely stable.
Rigorous proofs of these facts are presented later in the manuscript, in Remark~\ref{rem:SDEA-not-2port-passive} and Lemma~\ref{lem:absolute-stability-b22-needed}.

It is well-established in the literature that SEA under VSIC is not one-port passive while rendering pure springs with spring constants larger than the physical stiffness of SEA\@.
Furthermore, it has also been shown that SEA under VSIC with integral controllers cannot passively render any VE having a Voigt model~\cite{Tagliamonte2014b,FatihEmre2020}.
Since two-port passive system can stably couple with any passive terminations, these results serve as counter-examples proving that the SEA under VSIC cannot be two-port passive.

SDEA under VSIC inherits a version of the physical stiffness upper-bound as in the SEA under VSIC while rendering pure stiffness.
Furthermore, while SDEA under VSIC can passively render Voigt models thanks to the addition of damping element to its physical filter, passivity can be ensured for only a limited range of Voigt model parameters.

Along these lines, SDEA under VSIC is also not two-port passive.

\vspace{-3mm}
\subsection{Virtual Coupler for VSIC of SDEA} \label{subsec:virtual-coupler-design-for-sdea}

\begin{figure}[!t]
	\begin{subfigure}[b]{\columnwidth}
		\centering
		\includegraphics[width=.9\textwidth]{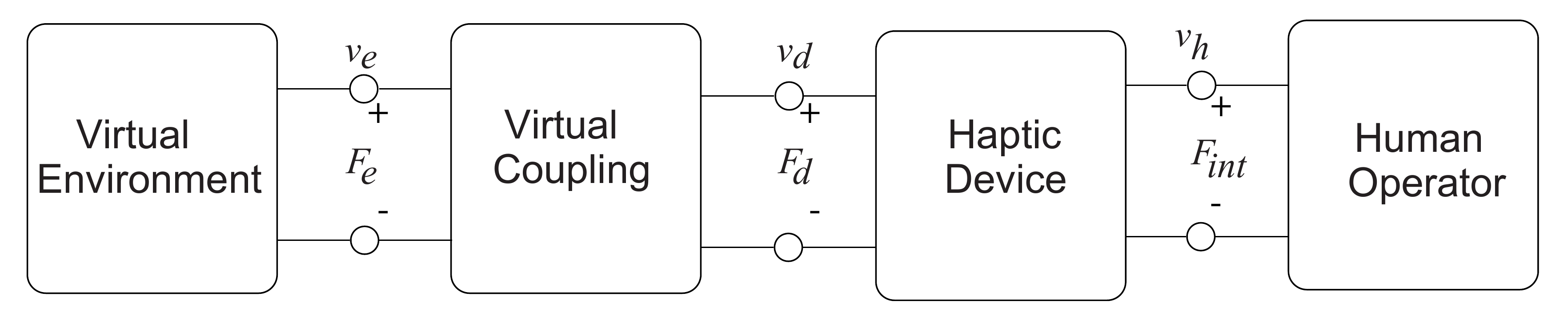}
		\vspace{-.5\baselineskip}
		\caption{Two-port network representation of SDEA under VSIC with a virtual coupler} \label{fig:twoportVC}
	\end{subfigure}
	\vspace{.05\baselineskip}
	\begin{subfigure}[b]{\columnwidth}
		\centering
		\includegraphics[width=1.0\textwidth]{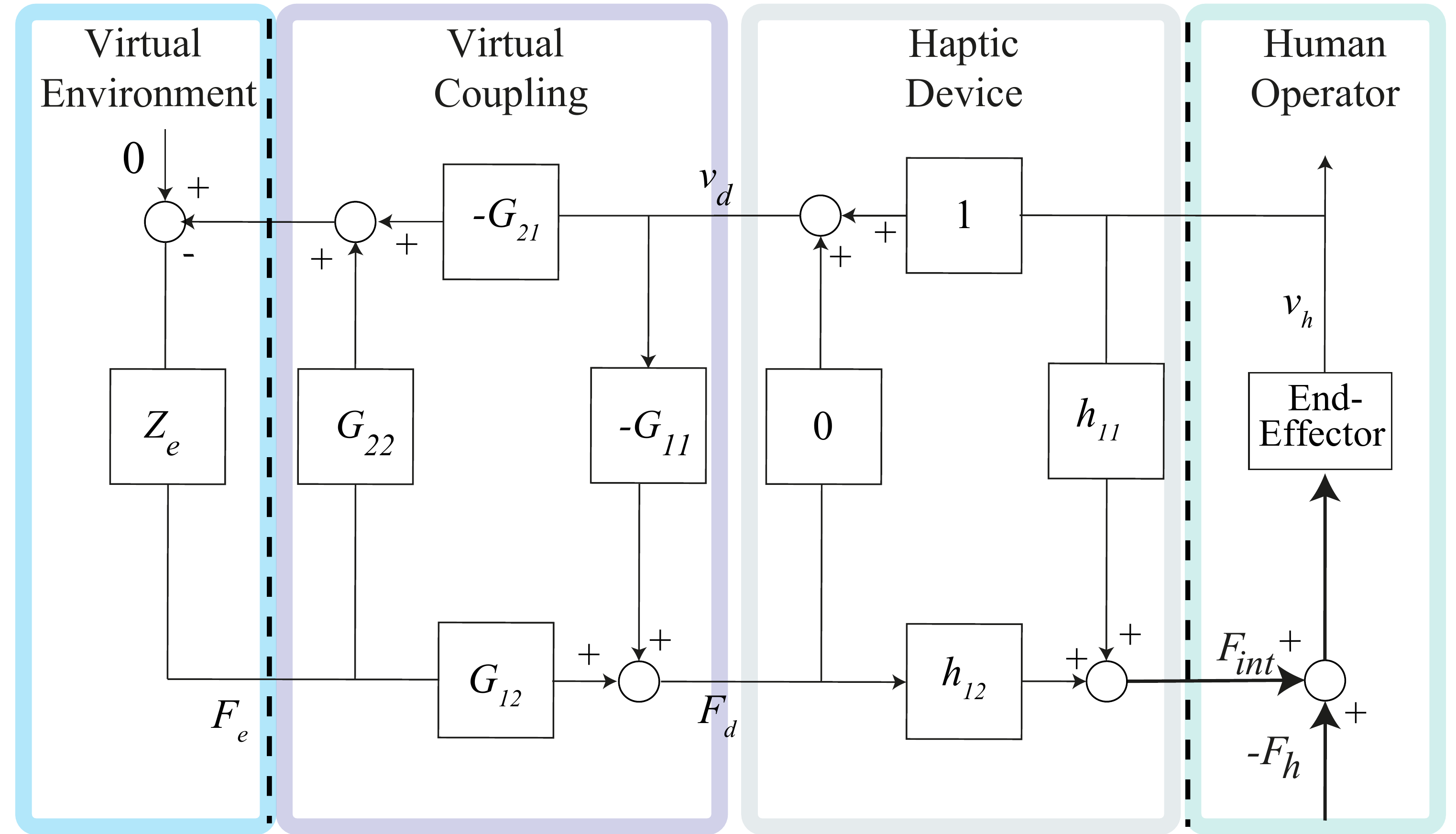}
		\vspace{-.5\baselineskip}
		\caption{Block diagram in Figure~\ref{fig:generalizedSEA} with a generic virtual coupler} \label{fig:generalizedVC}
	\end{subfigure}
	\vspace{-1\baselineskip}
	\caption{The network diagram and the corresponding block diagram of SDEA under VSIC with a virtual coupler} \label{fig:generalizedVC3}
	\vspace{-.3\baselineskip}
\end{figure}

Since the two-port model of SDEA under VSIC is not two-port passive, a virtual coupler (VC) is introduced before the VE, as suggested in the haptics literature~\cite{Colgate95, Adams99}.
Figure~\ref{fig:generalizedVC3} presents the network diagram and the corresponding block diagram of the system with a VC\@.
The hybrid matrix for the SDEA with a generic virtual coupler reads as:
\begin{equation}
	\label{eq:H_VC}
	\begin{bmatrix}
		F_{\text{int}} \\ v_e
	\end{bmatrix}
	=
	\begin{bmatrix}
		h_{11} & h_{12}    \\
		G_{21} & G_{22}
	\end{bmatrix}
	\begin{bmatrix}
		-v_h \\    F_e
	\end{bmatrix},
\end{equation}
where
\begin{equation*}
	h_{11} = \frac{(K_f + B_f s)(C_m P (1 + C_f G_{11}) + 1)}{(C_m P + 1) s + P (K_f + B_f s)(C_m C_f + 1)}
\end{equation*}

\begin{equation*}
	h_{12} = \frac{C_f C_m G_{12} P {(K_f + B_f s)}}{(C_m P + 1)s + P (K_f + B_f s)(C_m C_f + 1)}.
\end{equation*}

\begin{figure}[!t]
	\centering
	\resizebox{0.55\columnwidth}{!}
	{\rotatebox{0}{\includegraphics[width=\textwidth]{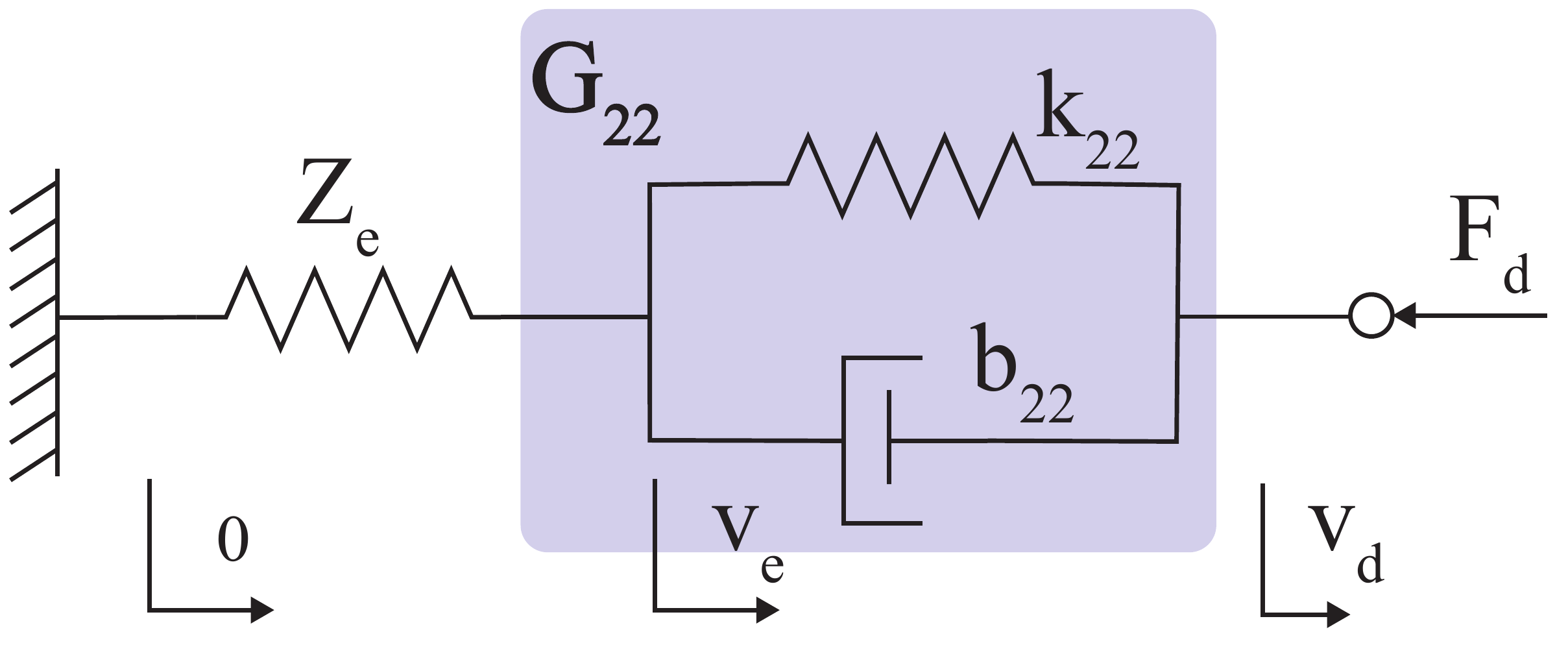}}}
	\vspace{-.3\baselineskip}
	\caption{Physical equivalent of the presented virtual coupler attached to a virtual environment (VE is depicted as a spring)} \label{fig:VC-physical}
	\vspace{-.5\baselineskip}
\end{figure}

Transfer functions $G_{ij}$ represent the two-port model of a generic VC\@.
The transfer function $G_{22}$, virtual environment, and the device are in series, as shown in Figure~\ref{fig:VC-physical}.
On the other hand, $G_{11}$ is parallel to this structure, coupling the ground of the VE and the device.
Transfer functions $G_{12}$ and $G_{21}$ represent the scaling factors between the forces and velocities, respectively.

Figure~\ref{fig:VC-physical} presents a virtual coupler form that is commonly used in the literature~\cite{Colgate95,Adams99,Griffiths2006}.
In this model, a physical equivalent for $G_{22}$ corresponds to a spring-damper pair ($k_{22}$ and $b_{22}$, respectively) in parallel.
Since haptic applications necessitate transferring the mechanical impedance of the VE transparently within stability regions, a natural selection for $G_{22}$ would be a stiff coupling.
At low frequencies, a stiff $k_{22}$ achieves this goal while, at high frequencies, $b_{22}$ compensates for the impedance drop of $k_{22}$.
Later, in Section~\ref{sec:TwoportPassivityAnalysis}, we justify this selection in terms of the coupled stability of the system.

Formally, the mathematical model of this VC used in our analysis is as follows:

\begin{equation}
	\label{eq:gselection}
	G_{11} = 0, \hspace{5mm} G_{12} = -G_{21} = 1, \hspace{5mm} G_{22} = \frac{s}{b_{22} s+k_{22}}.
\end{equation}

\vspace{-3mm}
\section{Two-Port Passivity of SDEA under VSIC} \label{sec:TwoportPassivityAnalysis}
In this section, we present the \emph{necessary and sufficient conditions} for two-port passivity for SDEA under VSIC.
To improve the readability of the section, we focus on the main results and present the proofs in the Appendix.

The hybrid matrix for the resulting two-port system can be expressed as:
\begin{equation}
	\label{eq:H-VC-full-order}
	\begin{bmatrix}
		F_{\text{int}} \\ v_e
	\end{bmatrix}
	=
	\begin{bmatrix}
		h_{11} & h_{12} \\
		-1     & h_{22}
	\end{bmatrix}
	\begin{bmatrix}
		-v_h \\ F_e
	\end{bmatrix},
\end{equation}
\noindent where
\begin{IEEEeqnarray}{rCl}
	\label{eq:generaltwoports}
	h_{11} &=&
	\frac{\splitdfrac{B_f M s^4 +\left(B_f(B + P_m) + K_f M\right) s^3}
	{+ \left(B_f I_m+ K_f(B + P_m)\right) s^2 + K_f I_m s}}{a_4 s^4 + a_3 s^3 + a_2 s^2+ a_1 s+ a_0}
	\IEEEyessubnumber\IEEEeqnarraynumspace\label{eq:h11}
\end{IEEEeqnarray}
\begin{IEEEeqnarray}{rCl}
	h_{12} &=&
	\frac{\splitdfrac{B_f P_m P_f s^3 + P_m P_f (K_f + B_f (\mu + \nu)) s^2}{ + (B_f I_m I_f + K_f P_m P_f (\mu + \nu)) s + K_f I_m I_f}}{a_4 s^4 + a_3 s^3 + a_2 s^2+ a_1 s+ a_0}\;\;\;\;\;\;\; \IEEEyessubnumber \label{eq:h12}\\
	h_{22} &=& \frac{s}{b_{22} s+k_{22}} \IEEEyessubnumber \label{eq:h22}
\end{IEEEeqnarray}
\noindent with
\begin{IEEEeqnarray}{rCl}
	a_4 &=& M \nonumber\\
	a_3 &=& B + P_m + B_f (\alpha + P_m P_f) \nonumber\\
	a_2 &=& I_m + K_f (\alpha + P_m P_f) + B_f P_m P_f (\mu + \nu)) \nonumber\\
	a_1 &=& B_f I_m I_f + K_f P_m P_f (\mu + \nu) \nonumber\\
	a_0 &=& K_f I_m I_f, \nonumber
\end{IEEEeqnarray}
\noindent and \vspace{-3mm}
\begin{IEEEeqnarray}{rCl}
	\mu & = & \frac{I_m}{P_m}, \;\;\; \nu = \frac{I_f}{P_f}. \nonumber
\end{IEEEeqnarray}
\noindent where $h_{11}$ and $h_{12}$ represent the system dynamics, and $h_{22}$ contains the terms of VC.

Following lemmas are instrumental in the derivation of the \emph{necessary and sufficient conditions} for two-port passivity for SDEA under VSIC\@.

\begin{lemma}[\cite{FatihEmre2020,Chen2009}]
	\label{lem:RH}
	Let $f(s) = a_4 s^4 + a_3 s^3 + a_2 s^2 + a_1 s + a_0$ for $a_i > 0$ be the characteristic equation of a fourth-order system.
	Then, $f(s)$ has no roots in the open right half plane \textbf{if and only if} $a_1(a_2 a_3 - a_1 a_4) - a_0 a_3^2 \geq 0 $.
\end{lemma}
\begin{IEEEproof}
	The proof has been presented in~\cite{FatihEmre2020}.
\end{IEEEproof}
\begin{lemma}
	\label{lem:residue}
	Given a real-rational function
	\begin{equation}
		\label{eq:Zs4}
		Z(s) = \frac{s\,(b_3 s^3 + b_2 s^2 + b_1 s + b_0)}{a_4 s^4 + a_3 s^3 + a_2 s^2 + a_1 s + a_0},
	\end{equation}
	where $a_i > 0$ and $b_i > 0$, $Z(s)$ has a simple, conjugate pair of poles on the imaginary axis \textbf{if and only if} $a_1 (a_2 a_3 - a_1 a_4) = a_0 a_3^2$.
\end{lemma}
\begin{IEEEproof}
	The proof is presented in the \hyperlink{pr:residue}{Appendix}.
\end{IEEEproof}
\begin{lemma}
	\label{lem:residue-pos-real}
	Consider the system in Eqn.~\eqref{eq:Zs4}, where $a_i > 0$, $b_i > 0$, and $a_1(a_2 a_3 - a_1 a_4) = a_0 a_3^2$.
	The residues of the pair of poles on the imaginary axis are positive and real \textbf{if and only if} both of the following conditions hold.
	\begin{enumerate}
		\item[(a)] $a_1 b_3 - a_3 b_1 = (a_3 b_0 - a_1 b_2) a_3^2/(a_2 a_3 - 2 a_1 a_4)$
		\item[(b)] $a_1 b_3 - a_3 b_1 < 0$
	\end{enumerate}
\end{lemma}
\begin{IEEEproof}
	The proof is presented in the \hyperlink{pr:residue-pos-real}{Appendix}.
\end{IEEEproof}
\begin{lemma}
	\label{lem:positive-thirdpoly}
	Let $p(x) = p_3 x^3 + p_2 x^2 + p_1 x+p_0$ be any real polynomial.
	Then, $p(x)\geq0$ for all $x\geq0$ \textbf{if and only if} $p_3 \geq 0$ and $p_0 \geq0$ and one of the following conditions holds:
	\begin{enumerate}[label=(\alph*)]
		\item $p_1 \geq 0$ \emph {and} $p_2 \geq -\sqrt{3 p_1 p_3}$
		\item $\sigma$ = $p_2^2 - 3 p_1 p_3  > 0$ \emph {and} $p_1 p_2 - 9p_0 p_3 < 0$ \emph {and}\\
		$4 p_2 (p_1 p_2 - 9 p_0 p_3) < 4 p_1 \sigma + 3 p_3 \dfrac{(p_1 p_2 - 9 p_0 p_3)^2}{\sigma}$
	\end{enumerate}
\end{lemma}
%
\begin{IEEEproof}
	Our proof is based on an application of Strum's theorem and is presented in the \hyperlink{pr:positive-thirdpoly}{Appendix}.
	An alternative geometric proof can be found in~\cite{Chen2009}.
\end{IEEEproof}

Utilizing Lemmas~\ref{lem:RH}--\ref{lem:positive-thirdpoly}, Theorem~\ref{thm:SDEA-2port-pass} presents the \emph{necessary and sufficient conditions} for two-port passivity for SDEA under VSIC.
\begin{theorem}
	\label{thm:SDEA-2port-pass}
	Consider SDEA under VSIC as in Eqn.~\eqref{eq:H-VC-full-order}, where $K_f, M, B, P_m, P_f, k_{22}$ are taken as positive, while $B_f, I_m, I_f, b_{22}$ are assumed to be non-negative.
	Then, this system is two-port passive \textbf{if and only if} Conditions (a)--(c) hold:
	\begin{enumerate}[label=(\alph*)]
		\item \hypertarget{thm:SDEA-2port-pass-a}{The h-parameters have no poles in the right half plane}
		\begin{IEEEeqnarray}{rl}
			\frac{K_f \mu \nu}{B_f \mu \nu + K_f (\mu + \nu)}( & B + P_m + B_f(\alpha +P_m P_f ))^2 \leq \hfill \nonumber\\
			B_f (\alpha + P_m P_f)[I_m & + K_f \alpha + P_m P_f (K_f + B_f (\mu + \nu))] \nonumber\\
			+ \bigg[\kappa_3 +\frac{B_f P_m P_f}{K_f} & \bigg((B + P_m)(\mu + \nu) - M \mu \nu\bigg)\bigg] K_f \nonumber\\
			+ I_m (B + P_m) \nonumber
		\end{IEEEeqnarray}
		\item If $h_{11}$, given in Eqn.~\eqref{eq:h11}, has a pair of poles on the imaginary axis, their residues are real and positive
		\begin{enumerate}[label=(\arabic*), itemsep=3pt]
			\item $0 < \beta = a_3 \big( K_f (B + P_m) + B_f I_m \big) - a_1 (B_f M)$ \emph{and}
			\item $\beta (a_2 a_3 - 2 a_1 a_4) = \big(a_3 (K_f I_m) - a_1 \big( B_f (B + P_m) + K_f M \big)\big) a_3^2$
		\end{enumerate}
		\item The system parameters simultaneously satisfy the Conditions (i) and (ii):
		\begin{enumerate}[label=(\roman*)]
			\item \hypertarget{thm:SDEA-2port-pass-c-i}{Condition (i1) or (i2) holds}:
			\begin{IEEEeqnarray}{rcl}
				(i1) \;\;\; 0 &\leq r_1 =& K_f^2 \kappa_3 + B_f I_m^2 + B_f^2 I_m \kappa_1 \nonumber\\
				&\textrm{\emph{and}}& \nonumber\\
				0 &\leq B_f \big(( & B + P_m)^2 + B_f \kappa_3 - 2 I_m M\big) \nonumber\\
				&& + M\sqrt{3 B_f r_1}\nonumber\\
				(i2) \;\;\; 0 &\;< \rho_1 =\;& B_f^2 \left((B + P_m)^2 + B_f \kappa_3 - 2 I_m M\right)^2 \nonumber\\
				&&-3 B_f M^2 r_1 \nonumber\\
				&\textrm{\emph{and}}& \nonumber\\
				0 &\;> \rho_2 =\;& r_1 B_f \left((B + P_m)^2 + B_f \kappa_3 - 2 I_m M\right) \nonumber\\
				&&  -9 B_f I_m M^2 K_f^2 \kappa_1 \nonumber\\
				&\textrm{\emph{and}}& \nonumber\\
				4 \rho_2 & B_f \big((B &+ P_m)^2 + B_f \kappa_3 - 2 I_m M\big) < \nonumber\\
				&& 4 r_1 \rho_1 + \frac{3 B_f M^2 \rho_2^2}{\rho_1} \nonumber
			\end{IEEEeqnarray}
			\item \hypertarget{thm:SDEA-2port-pass-c-ii}{$k_{22}^2 \leq 4 b_{22} I_m (I_m P_f - B I_f) \left(\dfrac{K_f}{I_m + \alpha K_f}\right)^2$ and\\ $0 < b_{22} \leq 4 B_f$ and Condition (ii1) or (ii
2) holds}:
			\begin{IEEEeqnarray}{rcl}
				(ii1) \;\;\; 0 &\leq t_1 =& 4 b_{22} r_1 + k_{22}^2 \tau_2 - b_{22}^2 (I_m + \alpha K_f)^2 \nonumber\\
				&\textrm{\emph{and}}& \nonumber \\
				0 &\leq 4 b_{22} & r_2 + b_{22}^2 \tau_2 - k_{22}^2 M^2 + M\sqrt{3 b_{22}} \nonumber\\
				(ii2) \;\;\; 0 &\;< \tau_3 =\;& \left(4 b_{22} r_2 + b_{22}^2 \tau_2 - k_{22}^2 M^2 \right)^2 \nonumber\\
				&& - 3 M^2 b_{22} \tau_1 t_1 \nonumber\\
				&\textrm{\emph{and}}& \nonumber \\
				0 &\;> \tau_4 =\;& t_1 (4 b_{22} r_2 + b_{22}^2 \tau_2 - k_{22}^2 M^2) \nonumber\\
				&& -9 M^2 b_{22} \tau_1 t_0 \kappa_1 \nonumber \\
				&\textrm{\emph{and}}& \nonumber \\
				4 \tau_4 & (4 b_{22} r_2 & + b_{22}^2 \tau_2 - k_{22}^2 M^2) < \nonumber \\
				&& 4 t_1 \tau_3 + \frac{3 M^2 b_{22} \tau_1 \tau_4^2}{\tau_3} \nonumber
			\end{IEEEeqnarray}
		\end{enumerate}
	\end{enumerate} \vspace{-3mm}
	\noindent where \vspace{-3mm}
	\begin{IEEEeqnarray}{rcl}
		\kappa_1 &=& P_f I_m - B I_f \IEEEnonumber\\
		\kappa_2 &=& B + P_m - M (\mu + \nu) \IEEEnonumber\\
		\kappa_3 &=& \alpha (B + P_m) + P_m P_f \kappa_2 \IEEEnonumber\\
		\tau_1 &=& 4 B_f - b_{22} \IEEEnonumber\\
		\tau_2 &=& 2 M \left(I_m + \alpha K_f \right) -\left(B + P_m + \alpha B_f \right)^2\IEEEnonumber\\
		r_2 &=& B_f \big((B + P_m)^2 + B_f \kappa_3 - 2 I_m M\big). \IEEEnonumber
	\end{IEEEeqnarray}
\end{theorem}
\begin{IEEEproof}
	The \hyperlink{pr:thm-5}{proof} is presented in the Appendix.
\end{IEEEproof}
\smallskip
\begin{remark}
	\label{rem:SDEA-not-2port-passive}
	Two-port passivity necessitates SDEA, instead of SEA, and a damping element in the VC.
	In particular, according to \hyperlink{thm:SDEA-2port-pass-c-ii}{Condition \emph{(c-ii)}} of Theorem~\ref{thm:SDEA-2port-pass}, two-port passivity of SDEA under VSIC cannot be satisfied if $B_f = 0$ (i.e.\ there exist no physical damping as in the case of SEA) or $b_{22} = 0$ (i.e.\ VC does not incorporate a virtual damping).
	In this case, the highest-degree term of Eqn.~\eqref{eq:poly-two-port-case-cii} in the proof of Theorem~\ref{thm:SDEA-2port-pass} is reduced to
	\begin{equation*}
		B_f = 0 \implies p_2 = -b_{22}^2 M^2 < 0,
	\end{equation*}
	\noindent violating two-port passivity.
	For the second case, the fifth-degree term drops leaving the highest term as
	\begin{equation*}
		b_{22} = 0 \implies -k_{22}^2 M^2 < 0,
	\end{equation*}
	\noindent resulting in a similar violation.
\end{remark}
\smallskip
\begin{remark}
	Note that, we made use of Lemmas~\ref{lem:ReHs} and~\ref{lem:positive-thirdpoly} in the derivation of the necessary and sufficient conditions of two-port passivity of the system.
	However, it is possible to obtain simpler, but only sufficient conditions to ensure passivity.
	In particular, if the system parameters are selected such that all coefficients of the polynomial given in Lemma~\ref{lem:positive-thirdpoly} are positive, then the polynomial is positive for all $x$.
	Then, \hyperlink{thm:SDEA-2port-pass-c-i}{Condition \emph{(c-i)}} of Theorem~\ref{thm:SDEA-2port-pass} is simplified to the following conditions.
	\begin{IEEEeqnarray}{rCl}
		\label{eq:full-order-sufficient-c-i}
		0 \leq r_0 &:& I_f \leq \dfrac{I_m P_f}{B} \IEEEyesnumber\IEEEyessubnumber\\
		0 \leq r_1 &:& 0 \leq K_f^2 \kappa_3 + B_f I_m^2 + B_f^2 I_m \kappa_1 \IEEEyessubnumber\\
		0 \leq r_2 &:& 0\leq B_f \kappa_3 + (B + P_m)^2 - 2 I_m M. \IEEEyessubnumber
	\end{IEEEeqnarray}

	Similar considerations simplify \hyperlink{thm:SDEA-2port-pass-c-ii}{Condition \emph{(c-ii)}} to
	\begin{IEEEeqnarray}{rCl}
		\label{eq:full-order-sufficient-c-ii}
		0 \leq t_0 &:& k_{22}^2 \leq 4 b_{22} I_m (I_m P_f - B I_f) \left(\dfrac{K_f}{I_m + \alpha K_f}\right)^2 \IEEEyesnumber\IEEEyessubnumber\\
		0 \leq t_1 &:& 0 \leq 4 b_{22} r_1 + k_{22}^2 \tau_2 - b_{22}^2 (I_m + \alpha K_f)^2 \IEEEyessubnumber\\
		0 \leq t_2 &:& 0 \leq 4 b_{22} r_2 + b_{22}^2 \tau_2 - k_{22}^2 M^2 \IEEEyessubnumber\\
		0 \leq t_3 &:& 0 < b_{22} \leq 4 B_f. \IEEEyessubnumber
	\end{IEEEeqnarray}

	Then, \emph{sufficient} conditions to ensure two-port passivity of the system given by Eqn.~\eqref{eq:H-VC-full-order} can be stated as Conditions \emph{(a)} and \emph{(b)} of Theorem~\ref{thm:SDEA-2port-pass}, and Eqns.~\eqref{eq:full-order-sufficient-c-i} and~\eqref{eq:full-order-sufficient-c-ii}. These equations form a set of explicit solutions of the virtual coupler elements (i.e., $k_{22}$ and $b_{22}$), which is not available in Theorem~\ref{thm:SDEA-2port-pass}.
\end{remark}
\smallskip
\begin{remark}
	\label{rem:I_m-needed}
	The integral gain $I_m$ of the motion controller is \emph{necessary} so that the virtual coupler may have a non-zero stiffness $k_{22}$ when the integral gain $I_f$ of the force controller is non-zero.
	In particular, it follows from the first condition on $k_{22}$, Condition~\hyperlink{thm:SDEA-2port-pass-c-ii}{\emph{(c-ii)}} of Theorem~\ref{thm:SDEA-2port-pass}, that if $I_m = 0$, then $k_{22} = 0$, leaving the virtual coupler with only a pure damping term $b_{22}$.
	Note that result is in good agreement with the one-port stiffness rendering analysis~\cite{FatihEmre2020}.
	However, when both integral gains are zero, it is both possible to render a virtual spring and increase the bound on it.
\end{remark}
\smallskip
\begin{remark}
	\label{rem:alpha-effect}
	In the analysis of the system, $(1 - \alpha)$ modulates the state-dependent feed-forward action.
	In the first condition on $k_{22}$, in Condition~\hyperlink{thm:SDEA-2port-pass-c-ii}{\emph{(c-ii)}} of Theorem~\ref{thm:SDEA-2port-pass}, if $\alpha = 0$, then $k_{22}^{\max}$ is increased.
	However, the other equations in Condition \emph{(c-ii)} have inverse behavior with this result.
	Overall, completely canceling the physical interaction force affects passivity adversely.
	Although it is hard to follow this result through the analytical expressions of Theorem~\ref{thm:SDEA-2port-pass}, a numerical analysis reveals that there is an optimal value for $\alpha$, as discussed in Section~\ref{subsec:numerical-passivity-analysis}.
\end{remark}
\smallskip
\begin{remark}
	\label{rem:2port-pass-a-ci-equal-1port-pass-null-space}
	In general, Conditions \emph{(a)--(c-i)} of Theorem~\ref{thm:2port-pass} (excluding the conditions on $h_{22}$) are equivalent to those of Theorem~\ref{thm:1port-pass}, and lead to one-port passivity of a system described by $h_{11}$ coupled to a null environment.
	Therefore, Conditions \emph{(a)--(c-i)} presented in Theorem~\ref{thm:SDEA-2port-pass} generalize one-port passivity results presented in~\cite{FatihEmre2020} for SEA under VSIC\@.
	In this equation, if $B_f=0$, one can recover the necessary and sufficient conditions for passively rendering null impedance using the SEA under VSIC~\cite{FatihEmre2020}.
\end{remark}
\smallskip

\begin{remark}
	\label{rem:absolute-stability}
	The necessary and sufficient conditions for two-port passivity presented in Theorem~\ref{thm:SDEA-2port-pass} can be relaxed by studying absolute stability\footnote{The MATLAB code is available in the public GitHub repository https://github.com/ugurmengilli/SEA-2port-analysis.} given in Theorem~\ref{thm:abs-stability}.
	Although the equations are hard to interpret, they are useful for numerical implementation.
	Numerical comparisons between two-port passivity and absolute stability are presented in Section~\ref{sec:Numerical-Evaluations}.
\end{remark}

\begin{lemma}
	\label{lem:absolute-stability-b22-needed}
	Consider SDEA under VSIC as in Eqn.~\eqref{eq:H-VC-full-order}, where $K_f, M, B, P_m, P_f, k_{22}$ are taken as positive, while $I_m, I_f$ are assumed to be non-negative.
	Let $B_f$ be positive and let Conditions \emph{(a)}--\emph{(c-i)} of Theorem~\ref{thm:SDEA-2port-pass} are satisfied\footnote{Note that, Conditions \emph{(a)}--\emph{(c-i)} of two-port passivity (i.e., Theorem~\ref{thm:2port-pass}) and absolute stability (i.e., Theorem~\ref{thm:abs-stability}) are identical in this case.}.
	Then the two-port model of the system \emph{can~not} be absolutely stable unless it incorporates a virtual coupler with some damping (i.e., $b_{22} > 0$).
\end{lemma} \vspace{-1mm}
\begin{IEEEproof}
	Note that the Conditions \emph{(a)}--\emph{(c-i)} of Theorem~\ref{thm:SDEA-2port-pass} are equivalent to those of Theorem~\ref{thm:abs-stability}, since $h_{22}$ is already passive.
	Following Theorem~\ref{thm:abs-stability} for $b_{22} = 0 $, Condition (\emph{c-ii}) can be rewritten as follows. \vspace{-1mm}
	\begin{IEEEeqnarray}{c}
		-\text{Re}(h_{12} h_{21}) - \abs{h_{12} h_{21}} \geq 0.
	\end{IEEEeqnarray}

	\noindent We can further simplify this equation by setting $h_{21} = -1$: \vspace{-1mm}
	\begin{IEEEeqnarray}{c}
		\text{Re}(h_{12})\geq  \abs{h_{12}} = \sqrt{\text{Re}(h_{12})^2 + \text{Im}(h_{12})^2},
	\end{IEEEeqnarray}
	\noindent where the only possibility is that the system dynamics comprise pure damping.
	For any realistic design, it is not possible to manufacture the system without any mass or compliance in the system.
	Hence, in the virtual coupler, the damping element is required for satisfying absolute stability.
\end{IEEEproof}

\begin{lemma}
	Consider SDEA under VSIC as in Eqn.~\eqref{eq:H-VC-full-order}, where $K_f, M, B, P_m, P_f, k_{22}$ are taken as positive, while $I_m, I_f$ are assumed to be non-negative.
	Let $b_{22}$ be positive and let Conditions \emph{(a)}--\emph{(c-i)} of Theorem~\ref{thm:SDEA-2port-pass} are already satisfied\footnotemark[\value{footnote}].
	Then, the two-port model of the system \emph{can not} be absolutely stable unless it incorporates a physical coupler with a parallel a damping (i.e.\ $B_f > 0$).
\end{lemma}
\begin{IEEEproof}
	Note that, the absence of $B_f$ leads to SEA\@.
	Therefore, Conditions \emph{(a)}--\emph{(c-i)} of Theorem~\ref{thm:SDEA-2port-pass} correspond to the necessary and sufficient conditions for one-port passivity of SEA rendering null space (see Remark~\ref{rem:2port-pass-a-ci-equal-1port-pass-null-space}), as presented in~\cite{FatihEmre2020}.
	When $B_f = 0$, \hyperlink{thm:SDEA-2port-pass-c-i}{Condition} \emph{(c-i-2)} of Theorem~\ref{thm:SDEA-2port-pass} becomes invalid and Condition \emph{(c-i-1)} is reduced to
	\begin{IEEEeqnarray}{rCl}
		\label{eq:abs-stability-SEA-c-i}
		0 \leq r_1 \implies M &\leq& \frac{{\left(\alpha + P_f P_m \right)}{\left(B+P_m \right)}}{P_m P_f (\mu + \nu)}. \IEEEyesnumber\IEEEnosubnumber
	\end{IEEEeqnarray}

	Following Condition \emph{(c-ii)} of Theorem~\ref{thm:abs-stability} and Lemma~\ref{lem:ReHs} lead to a polynomial inequality of the form $0 \leq x^2 (p_5 x^5 + p_4 x^4 + p_3 x^3 + p_2 x^2 + p_1 x + p_0)$.
	To ensure the positiveness of this high-degree polynomial, it is necessary that $p_5 \geq 0$ and $p_0 \geq 0$.
	However, we can deduce that $p_5 < 0$ under the condition given by inequality~\eqref{eq:abs-stability-SEA-c-i}:  \vspace{-2mm}
	\begin{IEEEeqnarray}{rCl}
		\label{eq:abs-stability-SEA-c-ii}
		0 & > & p_5 = -b_{22} P_m P_f \big(((P_m + B) - M (\mu + \nu) \big)^2 \\
		&& -4 M K_f \big((\alpha + P_f P_m)(P_m + B) - M P_m P_f (\mu + \nu) \big), \IEEEnonumber\IEEEnosubnumber
	\end{IEEEeqnarray}
	\noindent which concludes the proof.
\end{IEEEproof}

To summarize, in this section the \emph{necessary and sufficiency conditions} are derived for two-port passivity of SDEA under VSIC. The need for the damping element $B_f$ in SDEA is proven. Furthermore, it is shown that positive $b_{22}$ is necessary for two-port passivity and stiffness cannot be rendered if $I_m=0$.

\section{Performance Analysis of SDEA} \label{sec:PerformanceAnalysisofSDEA}

While the coupled stability of pHRI systems constitutes an imperative design criterion, the performance of the system is also significant for better behavior upon interactions.
Thus, we determine the analytical equations for the evaluation of the system performance via transparency and $Z_{\text{width}}$ concepts, as described in Section~\ref{subsec:prelim-performance}.

\subsection{Transparency of SDEA under VSIC} \label{subsec:performance-transparency}

The two-port analysis enables investigation of the performance for all passive terminations through the use of the transparency concept.
One can compare the h-matrix of the system to ideal transparency (given in Eqn.~\eqref{eq:trans}) to assess the frequency-dependent characteristics of transparency.
We have plotted each h-parameter for all frequencies in Section~\ref{sec:Numerical-Evaluations}  in an attempt to observe the behavior. Furthermore, it is also possible to investigate it analytically at low and high frequencies.

Using the Eqn~\eqref{eq:H-VC-full-order}, the h-matrix converges to the following form at high frequencies.
\begin{equation}
	\label{eq:trans1}
	\lim_{s\to\infty}
	\begin{bmatrix}
		F_{\text{int}} \\
		v_{e}
	\end{bmatrix}
	=
	\begin{bmatrix}
		B_f & 0\\
		-1 &  1/b_{22}
	\end{bmatrix}
	\begin{bmatrix}
		-v_h\\
		F_e
	\end{bmatrix}.
\end{equation}

It is desirable to minimize $B_f$ and maximize $b_{22}$ to achieve better transparency at high frequencies.
However, two-port passivity conditions impose an upper bound on $b_{22}$ that depends on $B_f$ which cannot be set to zero. Furthermore, ideal transparency is not achievable at high frequencies, as indicated by $h_{12}=0$.

Note that transparency may not be crucial for frequencies that are over the force-control bandwidth of the system, while safety is a concern for these frequency ranges.
This transparency analysis indicates that for a safe design, minimizing $B_f$ may help to decrease the magnitude of force transmitted to the operator at high frequencies. Note that if this is not feasible, the damper force may be mechanically limited, as proposed in~\cite{Ott2017}.

At low frequencies, $h$-matrix of the system converge to ideal transparency as:

\begin{equation}
	\label{eq:trans2}
	\lim_{s\to0}
	\begin{bmatrix}
		F_{\text{int}} \\
		v_{e}
	\end{bmatrix}
	=
	\begin{bmatrix}
		0 & 1\\
		-1 & 0 \\
	\end{bmatrix}
	\begin{bmatrix}
		-v_h\\
		F_e
	\end{bmatrix}.
\end{equation}

Thanks to the integral gain $I_f$ of the force controller, ideal transparency is achievable at low frequencies. As the frequency increases, the effect of $I_f$ diminishes, and the proportional gain $P_f$ prevails, as shown in Figure~\ref{fig:numeric-transparency}. Also, virtual stiffness $k_{22}$ dominates the virtual coupler behavior at low and medium frequencies.

\subsection{Z-Width of SDEA under VSIC} \label{subsec:performance-z-width}

Passively achievable impedance range, $Z_{\text{width}}$, of the system, together with the minimum transmitted impedance, $Z_{\min}$,  are also investigated. The minimum impedance $Z_{\min}$ for SDEA under VSIC can be computed as
\begin{equation}
	\label{eq:zmin}
	Z_{\min} = \frac{\splitdfrac{B_f M s^4 +[B_f(B + P_m) + K_f M] s^3}
	{+ [B_f I_m+ K_f(B + P_m)] s^2 + K_f I_m s}}{a_4 s^4 + a_3 s^3 + a_2 s^2+ a_1 s+ a_0},
\end{equation}
\normalsize
\noindent where $a_4 = M$ and $a_0=K_fI_mI_f$. At low and high frequencies,
\begin{IEEEeqnarray}{rcl}
	\lim_{s\to0} Z_{\min}&=& 0 \IEEEyesnumber\IEEEyessubnumber\\
	\lim_{s\to\infty} Z_{\min}&=& B_f. \IEEEyessubnumber
\end{IEEEeqnarray}

\noindent These limits recommend low physical damping, $B_f$, and high integral gain, $I_f$, of the force controller to achieve low $Z_{\min}$ values.

The VC in Figure~\ref{fig:VC-physical} does not affect the $Z_{\min}$ of the system. However, if a parallel compliance $G_{11}$ is employed, then this term increases the minimum impedance $Z_{\min}$.

Achievable impedance range $Z_{\text{width}}$ for SDEA under VSIC can be computed as
\small
\begin{equation}
	\label{eq:Zwidth}
	Z_{\text{width}} = \frac{\splitdfrac{(b_{22}B_f P_m P_f )s^4}{\splitdfrac{+\{b_{22}P_mP_f[B_f(\mu+\nu)+K_f] + k_{22}B_fP_mP_f\}s^3}{\splitdfrac{+\{k_{22}P_mP_f[B_f(\mu+\nu)+K_f]}{\splitdfrac{+b_{22}[K_fP_mP_f(\mu+\nu)+B_fI_mI_f]\}s^2}{\splitdfrac{+\{k_{22}[K_fP_mP_f(\mu+\nu)+B_fI_mI_f]}{+b_{22}K_fI_mI_f\}s + k_{22}K_fI_mI_f}}}}}}{a_4 s^5 + a_3 s^4 + a_2 s^3+ a_1s^2+a_0s},
\end{equation}
\normalsize
\noindent where $a_4 = M$ and $a_0=K_fI_mI_f$.

Evaluating Eqn.~\eqref{eq:Zwidth} at low and high frequencies,
\begin{IEEEeqnarray}{rcl}
	\lim_{s\to0}Z_{\text{width}}&\to&\infty \IEEEyesnumber\IEEEyessubnumber\\
	\lim_{s\to0}s Z_{\text{width}} &=& k_{22} \IEEEnonumber\IEEEyessubnumber\label{eq:Kwidth}\\
	\lim_{s\to \infty}Z_{\text{width}} &=& 0. \IEEEnonumber\IEEEyessubnumber
\end{IEEEeqnarray}

\noindent Eqn.~\eqref{eq:Kwidth} indicates that the stiffness transmitted to the operator is bounded at low frequencies by the stiffness of the VC.
Consistent with the transparency analysis, these results indicate that SDEA cannot render impedances at high frequencies.

In conclusion, poor rendering performance is expected at high frequencies since SDEA assumes the dynamics of its physical filter for frequencies that are over the force control bandwidth of the device.

\section{Numerical Evaluations} \label{sec:Numerical-Evaluations}

In this section, we investigate the effect of VC parameters on the two-port passivity, transparency, and transmitted impedance of the system.
In particular, passivity bounds derived in Section~\ref{sec:TwoportPassivityAnalysis} are studied through numerical simulations, considering the VC in Figure~\ref{fig:VC-physical}.
VC parameters $k_{22}$ and $b_{22}$ are studied, systematically, to analyze their individual effects on the system behavior.

Table~\ref{tab:system-parameters} presents the parameter values employed for the numerical simulations. The system parameters $J$, $B$, $B_f$, and $K_f$ are determined by system identification. The control parameters $P_m$ and $P_f$ are selected based on the physical actuator limits, and the integral gains are tuned such that the system exhibits a decent tracking performance.
We ensured that given these nominal parameters, the isolated system (i.e., $h_{11}$) is stable (according to Conditions \emph{(a)-(b)} of Theorem~\ref{thm:SDEA-2port-pass}), and positive real (according to Condition \emph{(c-i)} of Theorem~\ref{thm:SDEA-2port-pass}). In the next subsections, any improved parameter is selected within the constraints of Theorem~\ref{thm:SDEA-2port-pass}.

\begin{table}[h!]
	\centering
	\caption{System parameters used for the numerical analysis}
	\vspace{-.25\baselineskip}
	\label{tab:system-parameters}
	\resizebox{\columnwidth}{!}{
		\begin{tabular}{l l c l}
			\Xhline{2\arrayrulewidth}
			\textbf{Param.} & \textbf{Description}                       & \textbf{Nominal Value} & \textbf{Unit}        \\ \hline
			$K_{f}$            & Stiffness of SDEA                          & 362                    & $\unit{N.m/rad}$     \\
			$B_{f}$            & Damping of SDEA                            & 0.05                   & $\unit{N.m.s/rad}$   \\
			$J$                & Inertia of the actuator                    & 6.399 $10^{-4}$        & $\unit{kg.m}^2$      \\
			$B$                & Damping of the actuator                    & 0.169                  & $\unit{N.m.s/rad}$   \\
			$P_{m}$            & Proportional gain of the motion controller & 0.28                   & $\unit{N.m.s/rad}$   \\
			$I_{m}$            & Integral gain of the motion controller     & 100                    & $\unit{N.m/rad}$     \\
			$P_{f}$            & Proportional gain of the force controller  & 40                     & $\unit{rad/N.m.s}$   \\
			$I_{f}$            & Integral gain of the force controller      & 70                     & $\unit{rad/N.m.s}^2$ \\
			\Xhline{2\arrayrulewidth}
		\end{tabular}
	}
	\vspace{-1.75\baselineskip}
\end{table}

\begin{figure*}[b!]
	\centering
	\begin{subfigure}[b]{0.49\textwidth}
		\centering
		\includegraphics[width=\textwidth]{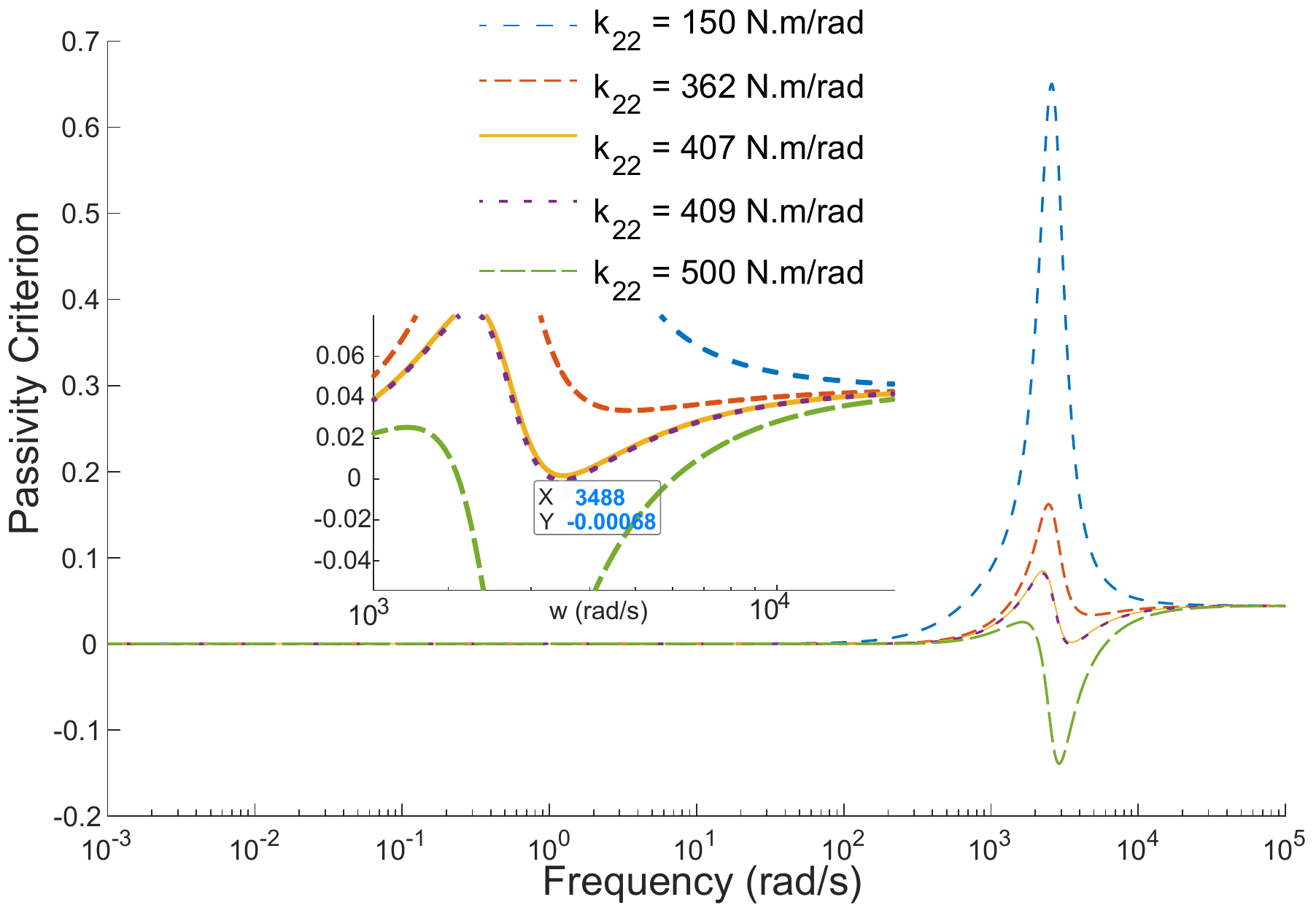}
		\vspace{-1.5\baselineskip}
		\caption{Effect of VC stiffness, $k_{22}$, on two-port passivity}
		\label{fig:k22effect}
	\end{subfigure}
	\hfill
	\begin{subfigure}[b]{0.49\textwidth}
		\centering
		\includegraphics[width=\textwidth]{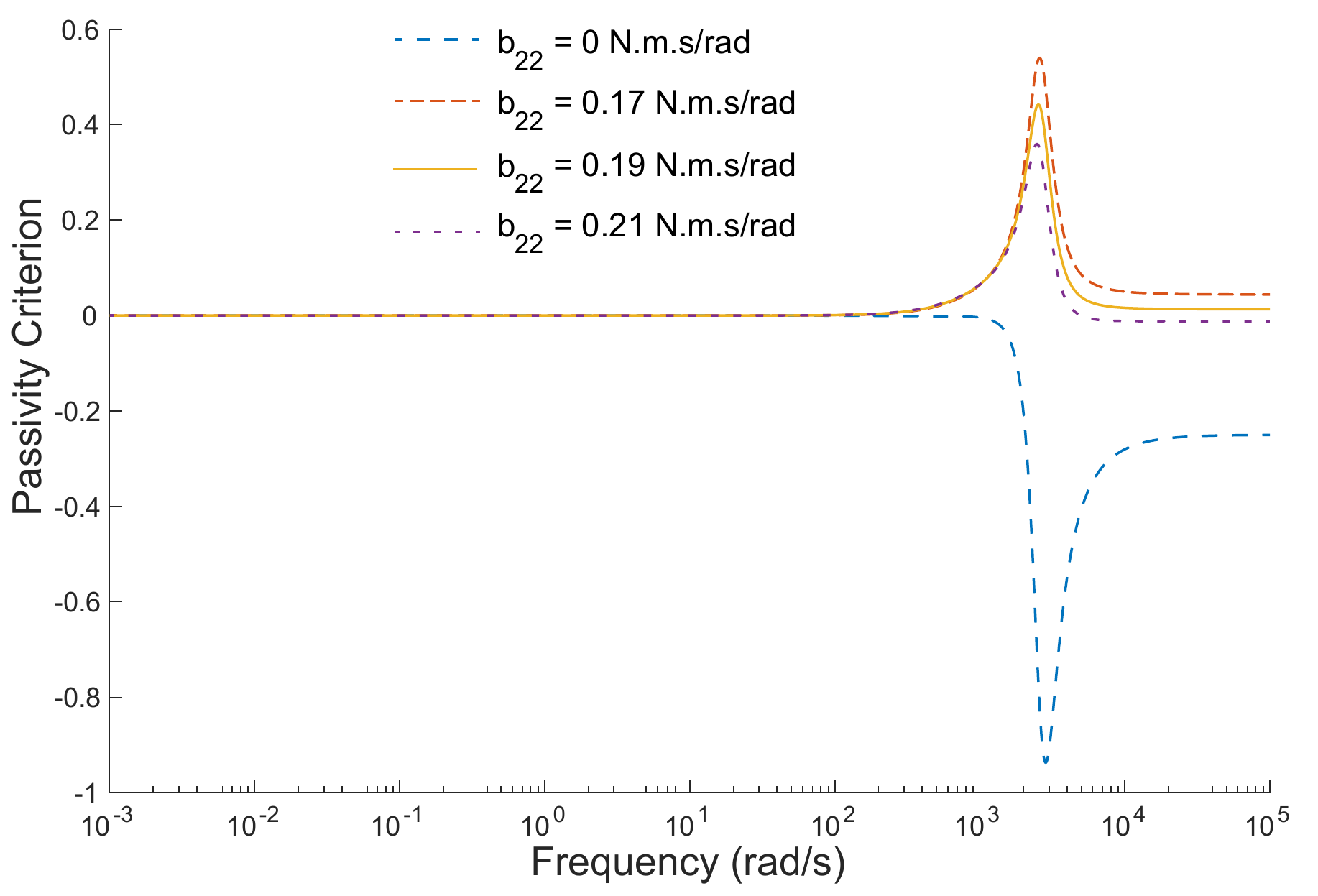}
		\vspace{-1.5\baselineskip}
		\caption{Effect of VC damping, $b_{22}$, on two-port passivity}
		\label{fig:b22effect}
	\end{subfigure}
	\vspace{-.25\baselineskip}
	\caption{Numerical evaluations of Condition \emph{(c-ii)} of Theorem~\ref{thm:2port-pass}}
	\label{fig:k22b22effect}
\end{figure*}

\subsection{Passivity Analysis} \label{subsec:numerical-passivity-analysis}

Figure~\ref{fig:k22b22effect} presents the effect of VC stiffness, $k_{22}$, and damping, $b_{22}$, on the system performance.
In these plots, passivity criterion corresponds to the evaluation of Condition \emph{(c-ii)} of Theorem~\ref{thm:2port-pass} (given as Eqn.~\eqref{eq:poly-two-port-case-cii} in the Appendix) according to the nominal values in Table~\ref{tab:system-parameters}.
Since parameters in the table already satisfy Conditions \emph{(a)--(c-i)}, each line in the plots must remain above zero for the two-port passivity.

Figure~\ref{fig:k22effect} reveals that a stiff virtual coupler adversely affects two-port passivity, causing the passivity criterion becomes negative.
The upper bound on $k_{22}$, as derived in Section~\ref{sec:TwoportPassivityAnalysis}, can also be observed in this plot.
For $k_{22} \gtrsim 408.5\unit{N.m/rad}$, the system becomes two-port active.
We also note that this value is greater than the physical stiffness of the SDEA in particular ($k_{22} \approx 1.17 K_f$), which any known SEA cannot passively render (see Section~\ref{subsubsec:spring-rendering} for a detailed discussion).

As presented in Section~\ref{sec:TwoportPassivityAnalysis}, VC damping $b_{22}$ is bounded by $4 B_f$.
Figure~\ref{fig:b22effect} verifies that the absence or even slightly overuse of $b_{22}$ makes the system two-port active.

Recall from Section~\ref{subsec:virtual-coupler-design-for-sdea} that $k_{22}$ is a concave function of $b_{22}$ when all other parameters are held constant (see Conditions \emph{(c-ii-1)} and \emph{(c-ii-2)} of Theorem~\ref{thm:SDEA-2port-pass}).
Therefore, it is possible to compute the $b_{22}$ value that maximizes $k_{22}$ for a given set of system parameters.
Table~\ref{tab:numeric-max-k22} lists the results of several numeric optimizations conducted for achieving the maximum $k_{22}$.

\begin{table}[b]
	\centering
	\caption{Maximum $k_{22}$ values for the full-order (FO) system with different feed-forward cancellation ratios compared to that of the FO system analyzed under absolute stability}
	\vspace{-.5\baselineskip}
	\label{tab:numeric-max-k22}
	\begin{tabular}{l c c}
		\Xhline{2\arrayrulewidth}
		\                                          & \textbf{Optimal} $b_{22}$ & \textbf{Max} $k_{22}$ \\
		\textbf{System Configuration}               & [N.m.s/rad]               & [N.m/rad]             \\ \hline
		With full feed-forward ($\alpha = 0$)       & 0.14                      & 367.0                 \\
		Without feed-forward ($\alpha = 1$)         & 0.17                      & 408.5                 \\
		With optimal feed-forward ($\alpha = 0.9) $ & 0.15                      & 415.5                 \\
		Without feed-forward (abs.\ stability)      & 0.13
		      & 432.0                 \\
		\Xhline{2\arrayrulewidth}
	\end{tabular}
	\vspace{-1\baselineskip}
\end{table}

In the analysis of the system, $(1 - \alpha)$ regulates the state-dependent feed-forward term.
Counter-intuitively, full cancelation of the physical interaction force adversely affects passivity, as observed in Table~\ref{tab:numeric-max-k22}.
Moreover, Conditions \emph{(c-ii-1)} and \emph{(c-ii-2)} of Theorem~\ref{thm:SDEA-2port-pass} include quadratic terms in $\alpha$ implying a concave behavior in $\alpha$ similar to the $b_{22}$ case.
Therefore, it is possible to maximize $k_{22}$ by selecting both $b_{22}$ and $\alpha$ optimally.
For instance a partial cancelation with $\alpha = 0.9$, $k_{22}$ can reach a maximum stiffness of $415.5\unit{N.m/rad}$.

For completeness, Table~\ref{tab:numeric-max-k22} provides the maximum achievable $k_{22}$ value according to Condition \emph{(c-ii)} of Theorem 2.
Absolute stability analysis can relax the passivity bounds on $k_{22}$ by 5\% compared to the two-port passivity of the system without the feed-forward cancelation.

In summary, the maximum achievable $k_{22}$ can be optimized via $b_{22}$ and $\alpha$, subject to the conditions of Theorem~\ref{thm:SDEA-2port-pass}.

\vspace{-3mm}

\subsection{Performance Analysis}\label{subsec:numerical-performance}

\begin{figure*}[ht!]
	\centering
	\begin{subfigure}[b]{0.49\textwidth}
		\centering
		\resizebox{.9750\columnwidth}{!}
		{\rotatebox{0}{\includegraphics[width=1\textwidth]{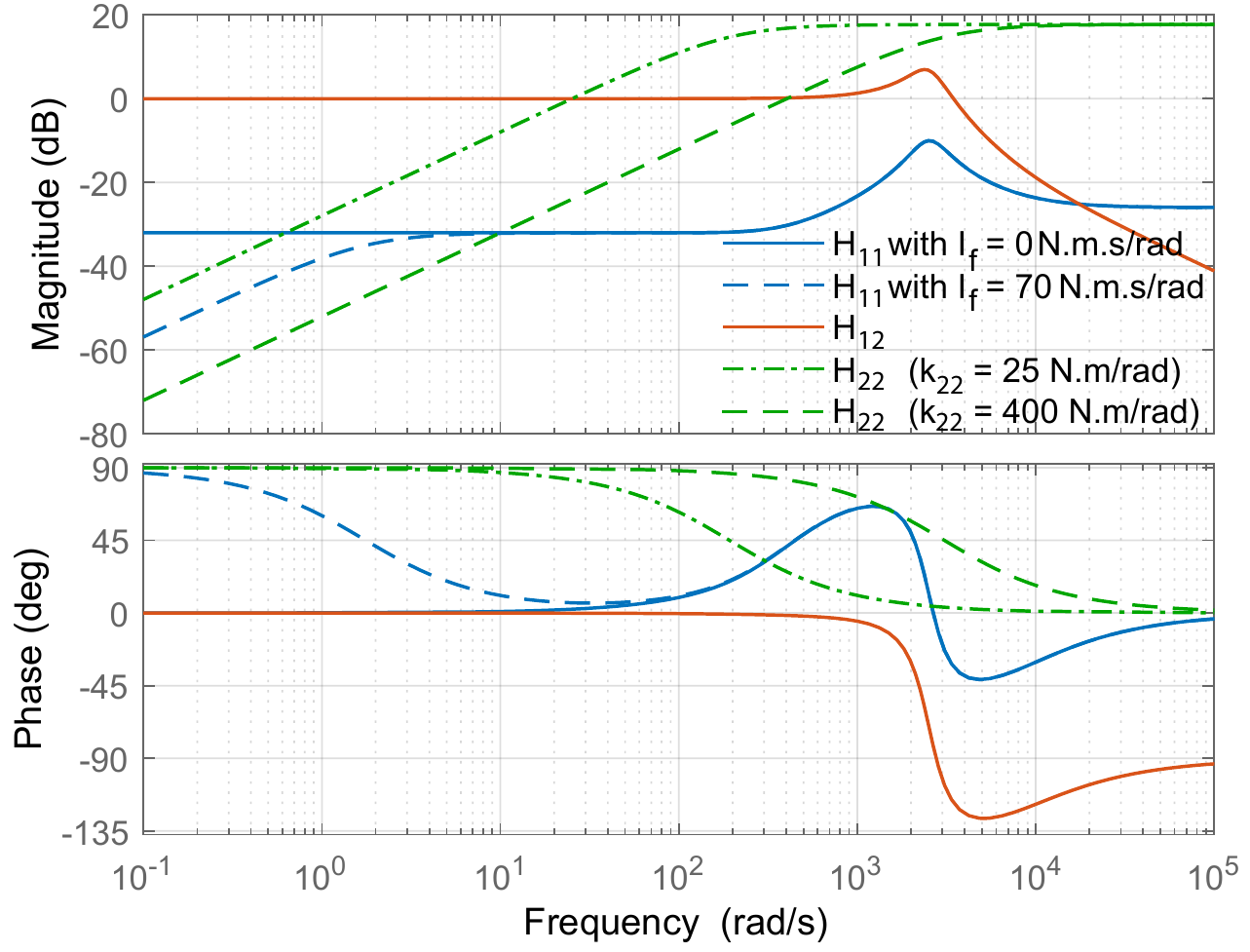}}}
		\vspace{-.25\baselineskip}
		\caption{The frequency response of the elements of the h-matrix to assess the transparency of the system. $h_{11}$ and $h_{22}$ are parasitic terms, and their magnitude plots are desired to be low. $h_{12}$ should be $0\unit{dB}$ ($\abs{h_{21}} = 0\unit{dB}$ and not shown in this plot).}
		\label{fig:numeric-transparency}
	\end{subfigure}
	\hfill
	\begin{subfigure}[b]{0.49\textwidth}
		\centering
		\resizebox{.9750\columnwidth}{!}
		{\rotatebox{0}{\includegraphics[width=1\textwidth]{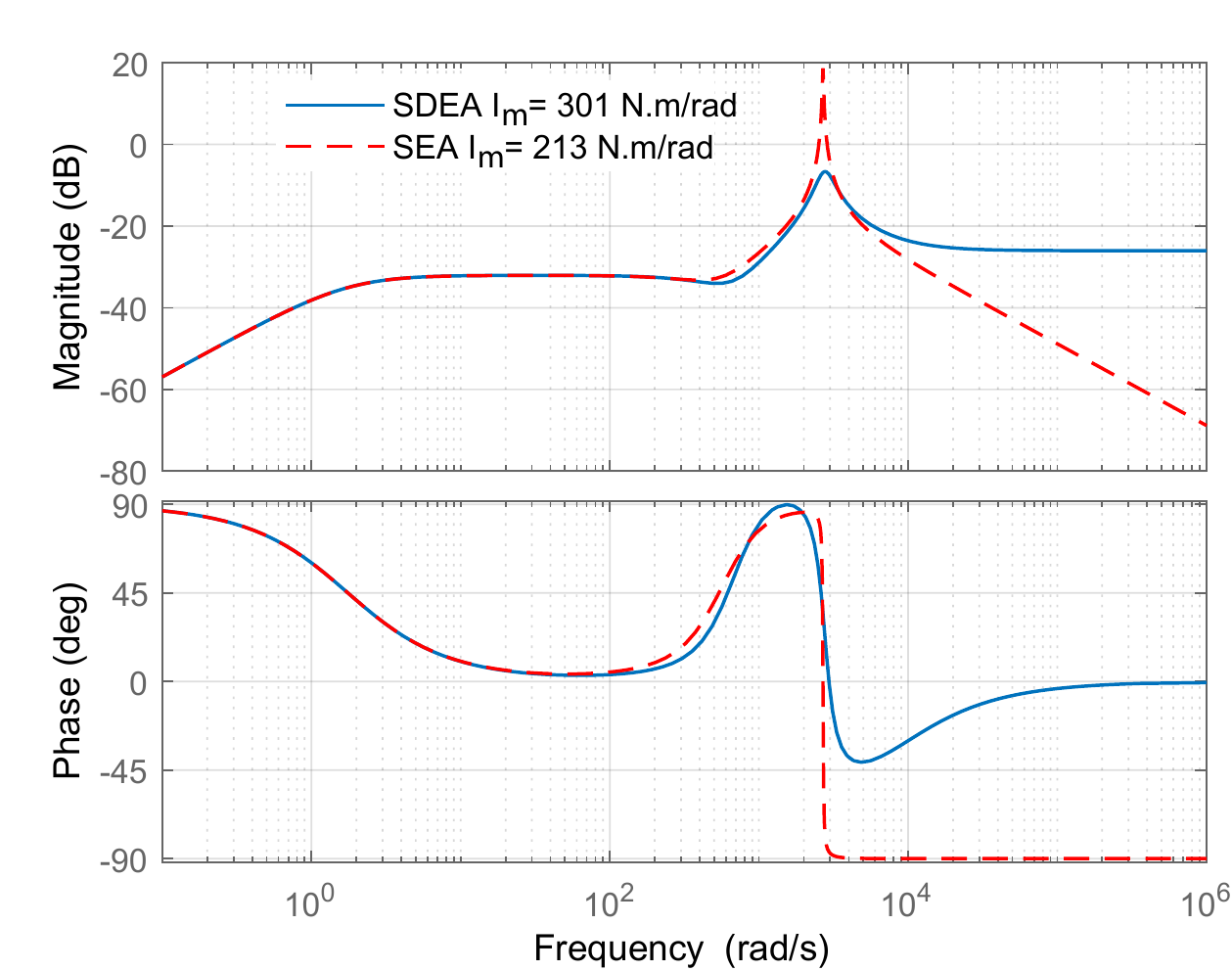}}}
		\vspace{-.25\baselineskip}
		\caption{Null impedance rendering of two-port passive SDEA compared to one-port passive SEA with maximum $I_m$ values. SDEA increases the phase margin around the resonance frequency and allows higher controller gains.}
		\label{fig:numeric-null-sdea-vs-sea}
	\end{subfigure} \\
	\begin{subfigure}[b]{0.49\textwidth}
		\centering
		\resizebox{.9750\columnwidth}{!}
		{\rotatebox{0}{\includegraphics[width=1\textwidth]{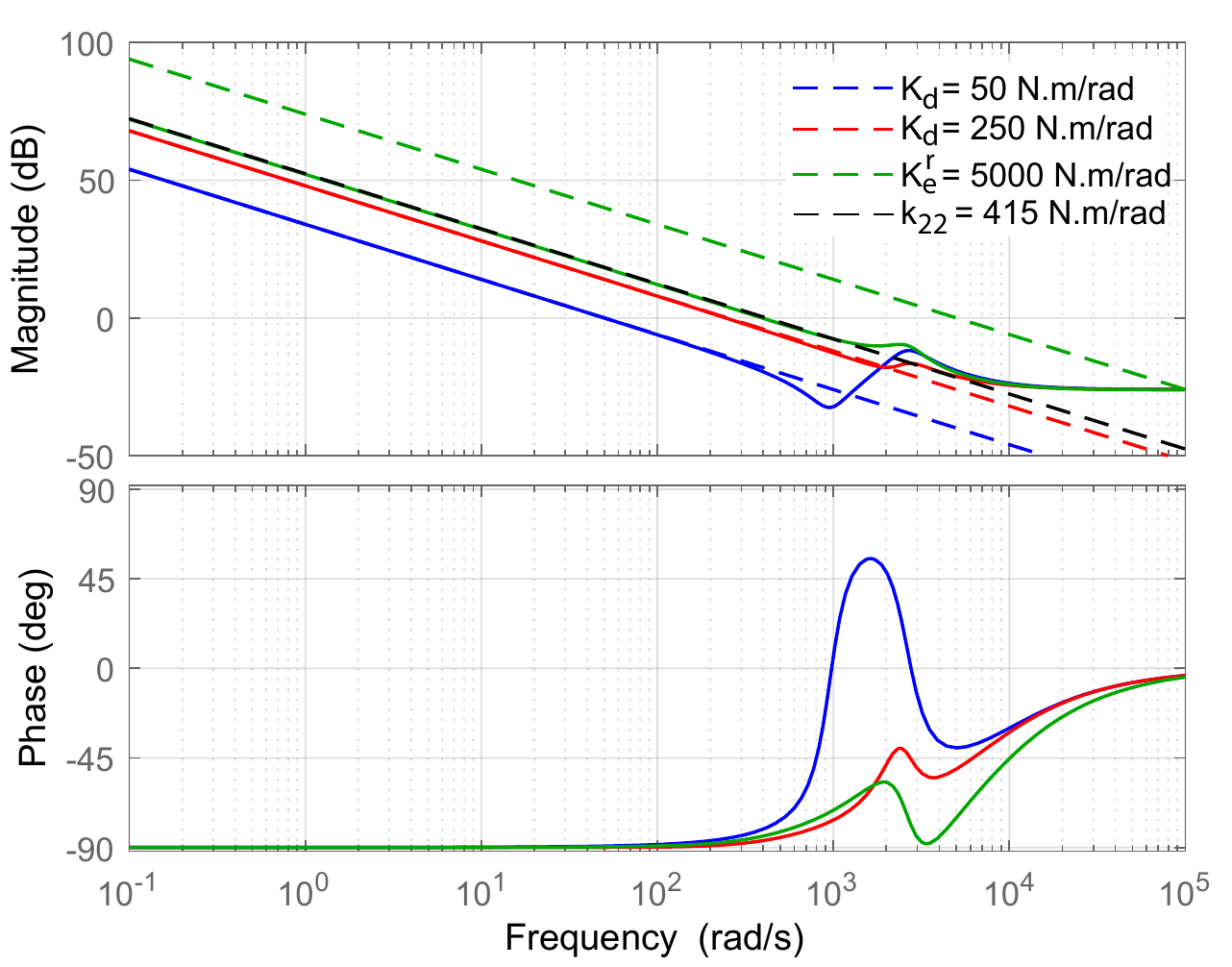}}}
        \vspace{-.25\baselineskip}
        \caption{Pure spring rendering of the SDEA (solid lines) compared to the desired stiffness (dashed lines) for various virtual stiffness values.\\}
		\label{fig:numeric-spring}
	\end{subfigure}
	\hfill
	\begin{subfigure}[b]{0.49\textwidth}
		\centering
		\resizebox{.9750\columnwidth}{!}
		{\rotatebox{0}{\includegraphics[width=1\textwidth]{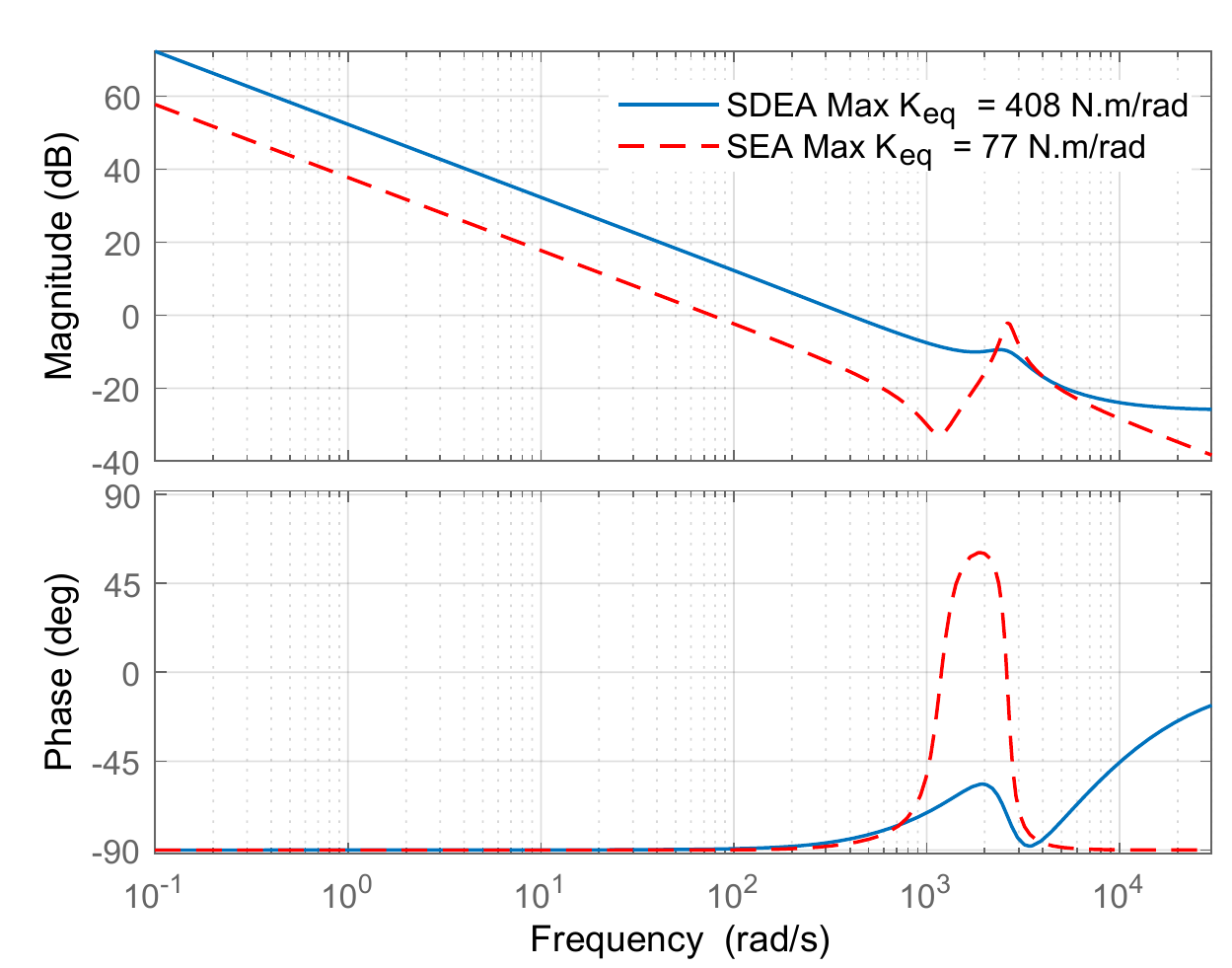}}}
		\vspace{-.25\baselineskip}
        \caption{Pure stiffness rendering of two-port passive SDEA compared to one-port passive SEA. SDEA increases the maximum virtual stiffness that can be rendered.}
		\label{fig:numeric-spring-sdea-vs-sea}
	\end{subfigure}
    \vspace{-.5\baselineskip}
	\caption{Performance of the two-port passive SDEA under VSIC}
	\vspace{-.75\baselineskip}
\end{figure*}

In Section~\ref{subsec:performance-transparency}, we have analytically studied the transparency of the system at the limit frequencies.
To observe the system behavior at intermediate frequencies, Figure~\ref{fig:numeric-transparency} plots the parameters of the \emph{h}-matrix.
Among these, $h_{11}$ and $h_{22}$ are the parasitic terms, and it can be observed that the transparency decreases as the frequency increases.

Increasing $I_f$ and $k_{22}$ improve the performance at low frequencies by decreasing parasitic effects due to $h_{11}$ and $h_{22}$, respectively.
However, $I_f$ possesses an upper bound due to Condition \emph{(c-i)} of Theorem~\ref{thm:SDEA-2port-pass}.
On the other hand, $I_m$ slightly reduces the mid-frequency magnitudes of $h_{11}$ while considerably enhancing $k_{22}^{\max}$.
In general, all proportional gains and damping terms (i.e., $B_f$ and $B$) smooth out and push the peaks of the plots to higher frequencies.
However, $B_f$ dominates the high-frequency response, distorting the transparency.

On the other hand, $\alpha$ affects neither magnitude nor phase of $h_{11}$ and $h_{12}$.
However, the optimal selection of $\alpha$ increases the $k_{22}^{\max}$, which improves the overall transparency of the system.

\subsubsection{Null Impedance Rendering}

In this subsection, we study null impedance rendering, i.e., $Z_e = 0$. In this case, $Z_{\text{to}}$ in Eqn.~\eqref{eq:zto} reduces to $h_{11}$.
Therefore, the following analysis appends to the comments on $h_{11}$ in the transparency analysis.

Figure~\ref{fig:numeric-null-sdea-vs-sea} compares the null impedance rendering performance of the investigated SDEA and the SEA in~\cite{FatihEmre2020} under nominal system parameters.
The phase plot of the figure exposes the improvement in the phase margin of the system due to the added parallel damping, which, in turn, grants increased bounds on the controller gains.
In particular, higher $I_m$ values can significantly improve the tracking performance and disturbance rejection of the inner motion control loop, such that the inner motion controller can act as an ideal motion source
within the control bandwidth. Furthermore, physical damping smooths the resonance peak that exists with SEA\@.

On the other hand, SDEA acts as a damper at high frequencies while SEA acts as a spring.
For the safety of interaction, it may be necessary to mechanically limit the interaction forces while utilizing SDEA\@.


\subsubsection{Spring Rendering} \label{subsubsec:spring-rendering}

In this subsection, we study the case of pure spring rendering, i.e., $Z_e = K_e / s$.
The impedance functions transmitted to the user under different virtual stiffness values are depicted in Figure~\ref{fig:numeric-spring} for the virtual coupler with $k_{22} = 415\unit{N.m/rad}$ and $b_{22} = 0.15\unit{N.m.s/rad}$.

Recall from Figure~\ref{fig:VC-physical} that the virtual environment comprises of the desired impedance and the VC\@.
Therefore, for a virtual spring and an ideally transparent device, the operator would feel an equivalent spring of stiffness $K_{eq} = \frac{k_{22} K_e}{k_{22} + K_e}$.
Then, it is possible to calculate the reference environment stiffness corresponding to the desired stiffness by solving the following equation for $K_{e}$.
\begin{equation}
	\label{eq:Ke-adjusted}
	K_e^r = \frac{k_{22} K_d}{k_{22} - K_d},
\end{equation}
\noindent where $K_e^r$ is the reference environment stiffness to render the desired stiffness, $K_d$.
However, since the environment is passive, $K_d < k_{22}$, noting that:
\begin{equation}
	\label{eq:Kd-limit}
	\lim_{K_d \to k_{22}} K_e^r \to \infty.
\end{equation}

Although two-port passivity allows all possible passive environments (even with unbounded parameters), Eqn.~\eqref{eq:Ke-adjusted} reveals that the VC practically limits the rendering performance.
Figure~\ref{fig:numeric-spring} verifies that, due to Eqn.~\eqref{eq:Ke-adjusted}, the SDEA can deliver the desired stiffness values below that of $k_{22}$ and saturates at $k_{22}$ for higher $K_e^r$ values (green dashed line in Figure~\ref{fig:numeric-spring}).

Figure~\ref{fig:numeric-spring-sdea-vs-sea} compares the performance of the SDEA with the SEA under identical system parameters.
Thanks to the physical damper, SDEA can passively render a virtual spring five times stiffer than that of SEA can passively render.

\begin{figure}[t]
	\centering
	\resizebox{.950\columnwidth}{!}
	{\rotatebox{0}{\includegraphics[width=1\textwidth]{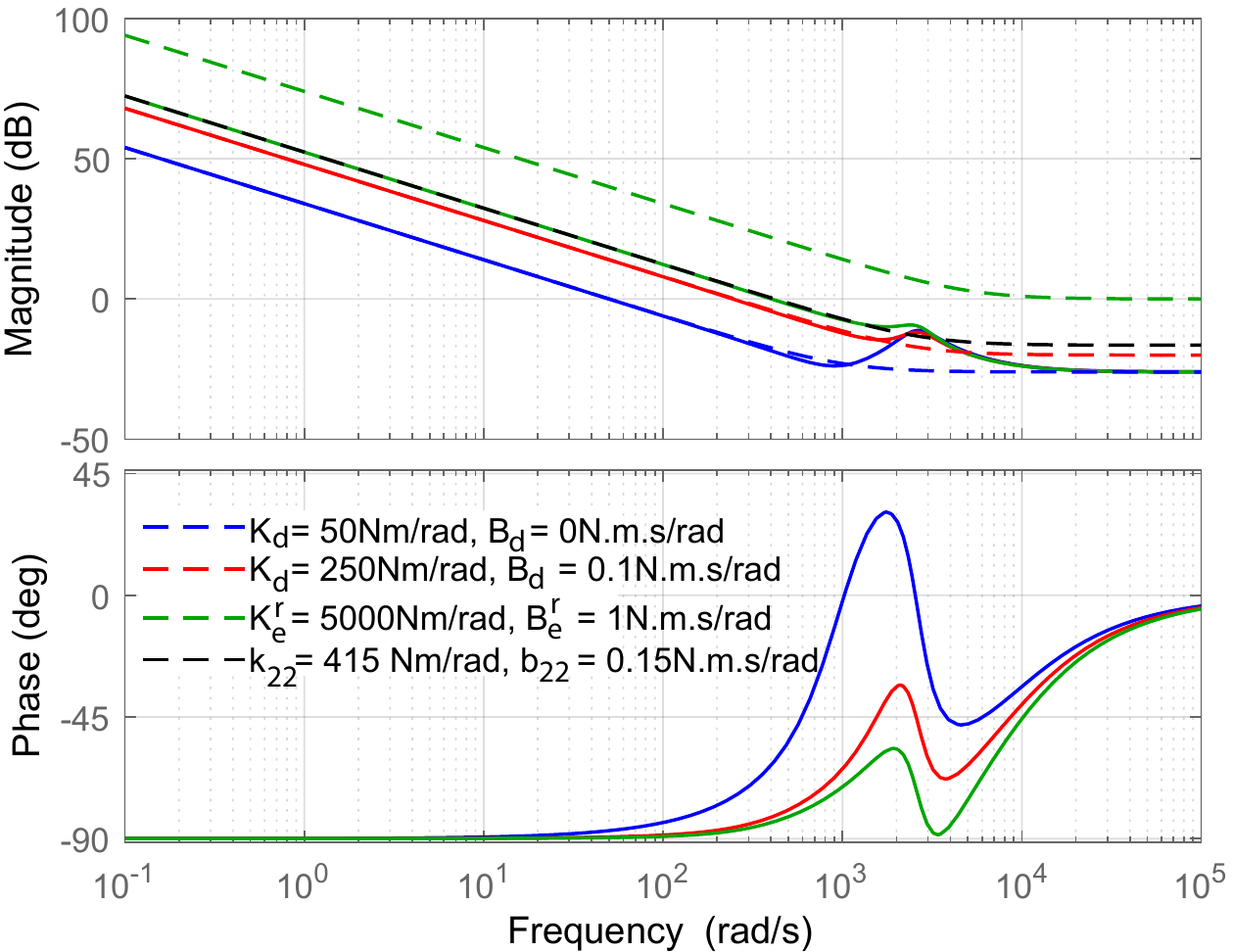}}}
	\vspace{-.5\baselineskip}
	\caption{Voigt model rendering of the SDEA (solid lines) with the desired stiffness and damping values (dashed lines).}
	\vspace{-.75\baselineskip}
	\label{fig:numeric-voigt}
\end{figure}

\smallskip

\subsubsection{Voigt Model Rendering}

In this subsection, we study Voigt model rendering, i.e., $Z_e = K_e / s + B_e$.
Figure~\ref{fig:numeric-voigt} presents the Voigt model rendering performance of SDEA, given different environment parameter selections.

Note that the equivalent stiffness model introduced in the previous subsection is also valid in this model.
\begin{equation}
	\label{eq:Be-adjusted}
	B_e^r = \frac{b_{22} B_d}{b_{22} - B_d},
\end{equation}
\noindent where $B_e^r$ is the reference environment damping to render the desired damping, $B_d$.
As in the case of pure stiffness rendering, the maximum virtual damping is also limited by the damping of the VC\@.
Similarly, the operator would feel the equivalent damping of the environment at high frequencies.

\section{Discussion} \label{sec:Discussion}

In this section, we review the results of the two-port passivity of SDEA under VSIC with VC and present general design guidelines.
We discuss the effects of the physical damping, virtual coupler, plant parameters, and controller gains in terms of the performance of haptic rendering within the two-port passivity limitations.

Independent of the application, it is generally a good practice to select high proportional gains that would not saturate the actuators within a reasonable range of frequencies~\cite{Tagliamonte2014b}, improving both stability and performance.
The following remarks present the trade-offs in the design procedure.

\vspace{-3mm}
\subsection{The Necessity of Physical Damping}

It has been well-established in the literature that SEA under VSIC cannot render the Voigt environment passively~\cite{Tagliamonte2014b}.
The inclusion of physical damping is crucial in that it enables SDEA to achieve Voigt model rendering while preserving one-port passivity.
Remark~\ref{rem:SDEA-not-2port-passive} in Section~\ref{sec:TwoportPassivityAnalysis} highlights that physical damping is also necessary to realize the two-port passivity of the device together with a VC\@.

The choice of the magnitude of $B_f$ affects the high-frequency response, as presented in Figures~\ref{fig:numeric-null-sdea-vs-sea},~\ref{fig:numeric-spring-sdea-vs-sea}, and~\ref{fig:numeric-voigt}.
In particular, the safety of interactions requires low $B_f$ to limit the magnitude of impact forces.
However, this also reduces the maximum $b_{22}$ that the VC can employ, which limits the maximum $k_{22}$ and the rendering performance.
Therefore, one possible design strategy would be to select the maximum $B_f$ that is acceptably safe for the application and iteratively trade-off the safety until the rendering performance becomes satisfactory.

\vspace{-3mm}
\subsection{The Necessity of Integral Gain of the Motion Controller}

Section~\ref{sec:TwoportPassivityAnalysis} proves that the integral gain $I_m$ of the motion controller is necessary for the virtual coupler of SDEA under VSIC to have a non-zero stiffness.
As discussed in Remark~\ref{rem:I_m-needed}, this result is in good agreement with the one-port passivity analysis in~\cite{FatihEmre2020} concluding that SEA cannot render a virtual spring when $I_m = 0$.

Note that the spring of the VC is not necessary for the passivity of the system while its magnitude sets an upper bound on the stiffness range that the device can display at low and intermediate frequencies.
Moreover, the first term in Condition in \hyperlink{thm:SDEA-2port-pass-c-ii}{\emph{(c-ii)}} of Theorem~\ref{thm:SDEA-2port-pass} implies that $I_f < I_m P_f / B$, requiring high values of $I_m$ for high~$I_f$.

\vspace{-3mm}
\subsection{The Effect of Virtual Coupler and System Dynamics on Two-Port Passivity and Transparency}\label{subsec:discussion-Effectofvirtualcoupler}

We have shown in Section~\ref{sec:TwoportPassivityAnalysis} that the damping $b_{22}$ of the VC must be positive for the two-port passivity of the system and has an optimal value for maximizing the stiffness $k_{22}$ of the VC, as captured by Conditions \emph{(c-ii)} of Theorem~\ref{thm:SDEA-2port-pass}.
Such an optimization is valuable if the controller gains and system parameters do not display large changes.

Transparency and $Z_{\text{width}}$ analyses indicate that $k_{22}$ should be selected as stiff as possible, as the maximum achievable impedance transferred from the environment to the operator is dominantly limited by $k_{22}$.
Especially, large $K_f$ and $I_m$ enhance $k_{22}^{\max}$.
Furthermore, null impedance rendering performance determines the minimum impedance $Z_{\min}$ of the system.
In particular, increasing $I_f$ and $I_m$ improve $Z_{\min}$ at low and intermediate-frequencies, respectively, as shown in Figure~\ref{fig:numeric-transparency}.
However, increasing $I_f$ decreases $k_{22}^{\max}$.

As the state-dependent feed-forward compensation increases (i.e., as $\alpha$ decreases), the overall damping of the system deteriorates because $\alpha$ almost always acts as a booster of $b_{22}$ in Condition \emph{(c-ii)} of Theorem~\ref{thm:SDEA-2port-pass}.
Therefore, the maximum value of $k_{22}$ also decreases.
In particular, VC stiffness, $k_{22}$, is a concave function of $\alpha$ when all other parameters are kept constant, as discussed in Section~\ref{sec:Numerical-Evaluations}.
As in the case of $b_{22}$, optimization of this parameter while keeping all other parameters constant may improve the rendering performance, as evidenced in Table~\ref{tab:numeric-max-k22}.

\vspace{-3mm}


\section{Conclusion} \label{sec:Conclusion}

We have provided the necessary and sufficient conditions for two-port passivity of SDEA under VSIC. Based on the newly established conditions, we have derived non-conservative passivity bounds for a virtual coupler. We have also proved the necessity of a physical damping term in parallel to the series elastic element to ensure two-port passivity (and absolute stability), even when a virtual coupler with a damping element is present. The physical damping element helps improve the control performance of the system, increasing the limits on the controller gains and the maximum stiffness of the virtual coupler. Furthermore, we have proved that, unlike SEA, SDEA can passively render virtual springs that are stiffer than the physical elastic element employed.

We have shown that feed-forward cancelation of the interaction force may deteriorate the upper limit on the stiffness of the virtual coupler.

Future works include an extension of these results to other control architectures and more general virtual coupler models.

\vspace{-3mm}
\section*{Acknowledgment}

This work has been partially supported by T\"{U}B\.{I}TAK Grant~\#216M200.

\vspace{-3mm}
\section*{Authors’ Contributions}

VP has conceived and designed the analysis and supervised the study. UM, UC and ZOO have made significant contributions to the two-port passivity analysis of the sufficient conditions of two-port passivity of SDEA. UM and UC have proposed the use of Sturm’s theorem for the determination of the positiveness of a symbolic polynomial and derived the necessary and sufficient conditions of two-port passivity of SDEA. All authors have contributed to the writing of the manuscript.

\small
\bibliographystyle{IEEEtran}
\bibliography{references}

\begin{thebibliography}{10}
\providecommand{\url}[1]{#1}
\csname url@rmstyle\endcsname
\providecommand{\newblock}{\relax}
\providecommand{\bibinfo}[2]{#2}
\providecommand\BIBentrySTDinterwordspacing{\spaceskip=0pt\relax}
\providecommand\BIBentryALTinterwordstretchfactor{4}
\providecommand\BIBentryALTinterwordspacing{\spaceskip=\fontdimen2\font plus
\BIBentryALTinterwordstretchfactor\fontdimen3\font minus
  \fontdimen4\font\relax}
\providecommand\BIBforeignlanguage[2]{{%
\expandafter\ifx\csname l@#1\endcsname\relax
\typeout{** WARNING: IEEEtran.bst: No hyphenation pattern has been}%
\typeout{** loaded for the language `#1'. Using the pattern for}%
\typeout{** the default language instead.}%
\else
\language=\csname l@#1\endcsname
\fi
#2}}

\bibitem{Hogan1985a}
N.~Hogan, ``{Impedance Control: An Approach to Manipulation: Part
  I---Theory},'' \emph{ASME Journal of Dynamic Systems, Measurement, and
  Control}, vol. 107, no.~1, pp. 1--7, 1985.

\bibitem{Colgate1988phd}
J.~Colgate, ``{The Control of Dynamically Interacting Systems},'' Ph.D.
  Dissertation, Massachusetts Institue of Technology, 1988.

\bibitem{Chae1987}
C.~H. An and J.~Hollerbach, ``Dynamic stability issues in force control of
  manipulators,'' in \emph{IEEE International Conference on Robotics and
  Automation}, vol.~4, 1987, pp. 890--896.

\bibitem{Eppinger1987}
S.~Eppinger and W.~Seering, ``Understanding bandwidth limitations in robot
  force control,'' in \emph{IEEE International Conference on Robotics and
  Automation}, vol.~4, 1987, pp. 904--909.

\bibitem{Howard90}
R.~D. Howard, ``Joint and actuator design for enhanced stability in robotic
  force control,'' Ph.D. dissertation, Massachusetts Institute of Technology,
  1990.

\bibitem{Pratt1995}
G.~Pratt and M.~Williamson, ``Series elastic actuators,'' in \emph{IEEE/RSJ
  International Conference on Intelligent Robots and Systems}, vol.~1, 1995,
  pp. 399--406.

\bibitem{Robinson1999}
D.~Robinson, J.~Pratt, D.~Paluska, and G.~Pratt, ``Series elastic actuator
  development for a biomimetic walking robot,'' in \emph{IEEE/ASME
  International Conference on Advanced Intelligent Mechatronics}, 1999, pp.
  561--568.

\bibitem{Sensinger2006b}
J.~W. {Sensinger} and R.~F. {ff. Weir}, ``Unconstrained impedance control using
  a compact series elastic actuator,'' in \emph{IEEE/ASME International
  Conference on Mechatronics and Embedded Systems and Applications}, 2006.

\bibitem{Sensinger2006}
------, ``Improvements to series elastic actuators,'' in \emph{2006 2nd
  IEEE/ASME International Conference on Mechatronics and Embedded Systems and
  Applications}, 2006, pp. 1--7.

\bibitem{Veneman2006}
J.~F. Veneman, R.~Ekkelenkamp, R.~Kruidhof, F.~C.~T. van~der Helm, and
  H.~van~der Kooij, ``A series elastic- and bowden-cable-based actuation system
  for use as torque actuator in exoskeleton-type robots,'' \emph{The
  International Journal of Robotics Research}, vol.~25, no.~3, pp. 261--281,
  2006.

\bibitem{Khatib2008}
M.~Zinn, O.~Khatib, B.~Roth, and J.~K. Salisbury, ``Large workspace haptic
  devices - a new actuation approach,'' \emph{Symposium on Haptic Interfaces
  for Virtual Environment and Teleoperator Systems}, 2008.

\bibitem{Kyoungchul2012}
K.~Kong, J.~Bae, and M.~Tomizuka, ``A compact rotary series elastic actuator
  for human assistive systems,'' \emph{IEEE/ASME Transactions on Mechatronics},
  vol.~17, no.~2, pp. 288--297, 2012.

\bibitem{Sarac2014}
M.~Sarac, M.~A. Ergin, A.~Erdogan, and V.~Patoglu, ``{AssistOn-Mobile: A}
  series elastic holonomic mobile platform for upper extremity
  rehabilitation,'' \emph{Robotica}, vol.~32, pp. 1433--1459, 2014.

\bibitem{Gillespie2014b}
R.~B. Gillespie, D.~Kim, J.~M. Suchoski, B.~Yu, and J.~D. Brown, ``Series
  elasticity for free free-space motion for free,'' in \emph{IEEE Haptics
  Symposium}, 2014, pp. 609--615.

\bibitem{Erdogan2016}
A.~Erdogan, B.~Celebi, A.~C. Satici, and V.~Patoglu, ``{AssistOn-Ankle: A}
  reconfigurable ankle exoskeleton with series-elastic actuation,''
  \emph{Autonomous Robots}, pp. 1--16, 2016.

\bibitem{Otaran2016}
A.~Otaran, O.~Tokatli, and V.~Patoglu, ``Hands-on learning with a series
  elastic educational robot,'' in \emph{Proceedings of the EuroHaptics
  (EuroHaptics 2016) as Lecture Notes in Computer Science}, 2016.

\bibitem{Munawar2016}
H.~Munawar and V.~Patoglu, ``{Gravity-Assist: A} series elastic body weight
  support system with inertia compensation,'' in \emph{IEEE/RSJ International
  Conference on Intelligent Robots and Systems}, 2016, pp. 3036--3041.

\bibitem{Woo2017}
H.~Woo, B.~Na, and K.~Kong, ``Design of a compact rotary series elastic
  actuator for improved actuation transparency and mechanical safety,''
  \emph{IEEE International Conference on Robotics and Automation}, 2017.

\bibitem{Caliskan2018}
U.~Caliskan, A.~Apaydin, A.~Otaran, and V.~Patoglu, ``A series elastic brake
  pedal to preserve conventional pedal feel under regenerative braking,'' in
  \emph{IEEE/RSJ International Conference on Intelligent Robots and Systems},
  2018, pp. 1367--1373.

\bibitem{Senturk2018}
Y.~M. Senturk and V.~Patoglu, ``{MRI-VisAct: A} bowden cable-driven mri
  compatible series viscoelastic actuator,'' \emph{Transactions of the
  Institute of Measurement and Control, Sage}, vol.~40, no.~8, pp. 2440--2453,
  2018.

\bibitem{Umut2020}
U.~Caliskan and V.~Patoglu, ``Efficacy of haptic pedal feel compensation on
  driving with regenerative braking,'' \emph{IEEE Transactions on Haptics},
  vol.~13, no.~1, pp. 175--182, 2020.

\bibitem{Kamadan2017}
A.~Kamadan, G.~Kiziltas, and V.~Patoglu, ``Co-design strategies for optimal
  variable stiffness actuation,'' \emph{IEEE/ASME Transactions on
  Mechatronics}, vol.~22, no.~6, pp. 2768--2779, 2017.

\bibitem{Kamadan2018}
------, ``A systematic design selection methodology for system-optimal
  compliant actuation,'' \emph{Robotica}, pp. 1--19, 2018.

\bibitem{Colgate1988}
J.~E. Colgate and N.~Hogan, ``Robust control of dynamically interacting
  systems,'' \emph{International Journal of Control}, vol.~48, no.~1, pp.
  65--88, 1988.

\bibitem{hannafordRyu}
B.~Hannaford and J.-H. Ryu, ``Time-domain passivity control of haptic
  interfaces,'' \emph{IEEE Transactions on Robotics and Automation}, vol.~18,
  no.~1, pp. 1--10, 2002.

\bibitem{ryuGeneral}
J.-H. Ryu, D.-S. Kwon, and B.~Hannaford, ``Stability guaranteed control: time
  domain passivity approach,'' \emph{IEEE Transactions on Control Systems
  Technology}, vol.~12, no.~6, pp. 860--868, 2004.

\bibitem{Buerger2007}
S.~P. Buerger and N.~Hogan, ``Complementary stability and loop shaping for
  improved human--robot interaction,'' \emph{IEEE Transactions on Robotics},
  vol.~23, no.~2, pp. 232--244, 2007.

\bibitem{Aydin2018}
Y.~Aydin, O.~Tokatli, V.~Patoglu, and C.~Basdogan, ``Stable physical
  human-robot interaction using fractional order admittance control,''
  \emph{IEEE Transactions on Haptics}, vol.~11, no.~3, pp. 464--475, 2018.

\bibitem{Haddadi2010}
A.~Haddadi and K.~Hashtrudi-Zaad, ``Bounded-impedance absolute stability of
  bilateral teleoperation control systems,'' \emph{IEEE Transactions on
  Haptics}, vol.~3, no.~1, pp. 15--27, 2010.

\bibitem{Willaert2011}
B.~{Willaert}, M.~{Franken}, H.~{Van Brussel}, and E.~B. {Vander Poorten}, ``On
  the use of shunt impedances versus bounded environment passivity for
  teleoperation systems,'' \emph{IEEE International Conference on Robotics and
  Automation}, pp. 2111--2117, 2011.

\bibitem{Lee2019}
H.~{Lee}, J.~{Lee}, J.~{Ryu}, and S.~{Oh}, ``Relaxing the conservatism of
  passivity condition for impedance controlled series elastic actuators,'' in
  \emph{IEEE/RSJ International Conference on Intelligent Robots and Systems},
  2019, pp. 7610--7615.

\bibitem{Wyeth2008}
G.~{Wyeth}, ``Demonstrating the safety and performance of a velocity sourced
  series elastic actuator,'' in \emph{IEEE International Conference on Robotics
  and Automation}, 2008, pp. 3642--3647.

\bibitem{Pratt2004}
G.~Pratt, P.~Willisson, C.~Bolton, and A.~Hofman, ``Late motor processing in
  low-impedance robots: impedance control of series-elastic actuators,''
  \emph{American Control Conference}, pp. 3245--3251, 2004.

\bibitem{FatihEmre2020}
F.~E. Tosun and V.~Patoglu, ``Necessary and sufficient conditions for the
  passivity of impedance rendering with velocity-sourced series elastic
  actuation,'' \emph{IEEE Transactions on Robotics}, vol.~36, no.~3, pp.
  757--772, 2020.

\bibitem{Vallery2007}
H.~{Vallery}, R.~{Ekkelenkamp}, H.~{van der Kooij}, and M.~{Buss}, ``Passive
  and accurate torque control of series elastic actuators,'' in \emph{IEEE/RSJ
  International Conference on Intelligent Robots and Systems}, 2007, pp.
  3534--3538.

\bibitem{Vallery2008}
H.~{Vallery}, J.~{Veneman}, E.~{van Asseldonk}, R.~{Ekkelenkamp}, M.~{Buss},
  and H.~{van Der Kooij}, ``Compliant actuation of rehabilitation robots,''
  \emph{IEEE Robotics Automation Magazine}, vol.~15, no.~3, pp. 60--69, 2008.

\bibitem{Tagliamonte2014b}
N.~L. Tagliamonte and D.~Accoto, ``Passivity constraints for the impedance
  control of series elastic actuators,'' \emph{The Institution of Mechanical
  Engineers, Part I: Journal of Systems and Control Engineering}, vol. 228,
  no.~3, pp. 138--153, 2014.

\bibitem{Fiorini2017}
A.~Calanca, R.~Muradore, and P.~Fiorini, ``{Impedance control of series elastic
  actuators: Passivity and acceleration-based control},'' \emph{Mechatronics},
  vol.~47, pp. 37--48, 2017.

\bibitem{Newman1992}
W.~S. Newman, ``{Stability and Performance Limits of Interaction
  Controllers},'' \emph{Journal of Dynamic Systems, Measurement, and Control},
  vol. 114, no.~4, pp. 563--570, 1992.

\bibitem{Dohring2002}
M.~Dohring and W.~S. Newman, ``{Admittance enhancement in force feedback of
  dynamic systems},'' in \emph{IEEE International Conference on Robotics and
  Automation}, vol.~1.\hskip 1em plus 0.5em minus 0.4em\relax IEEE, 2002, pp.
  638--643.

\bibitem{Chew2004}
C.-M. Chew, G.-S. Hong, and W.~Zhou, ``Series damper actuator: a novel
  force/torque control actuator,'' in \emph{IEEE/RAS International Conference
  on Humanoid Robots}, vol.~2, 2004, pp. 533--546.

\bibitem{Hurst2004}
J.~Hurst, A.~Rizzi, and D.~Hobbelen, ``Series elastic actuation: Potential and
  pitfalls,'' in \emph{International conference on climbing and walking
  robots}, 2004.

\bibitem{Oblak2011}
J.~Oblak and Z.~Matja{\v{c}}i{\'{c}}, ``Design of a series visco-elastic
  actuator for multi-purpose rehabilitation haptic device,'' \emph{Journal of
  NeuroEngineering and Rehabilitation}, vol.~8, no.~1, p.~3, 2011.

\bibitem{Garcia2011}
E.~Garcia, J.~Arevalo, G.~Mu{\~{n}}oz, and P.~Gonzalez-de Santos, ``{Combining
  series elastic actuation and magneto-rheological damping for the control of
  agile locomotion},'' \emph{Robotics and Autonomous Systems}, vol.~59, no.~10,
  pp. 827--839, 2011.

\bibitem{Laffranchi2011}
M.~Laffranchi, N.~Tsagarakis, and D.~Caldwell, ``{A compact compliant actuator
  (CompAct) with variable physical damping},'' in \emph{IEEE International
  Conference on Robotics and Automation}, 2011, pp. 4644--4650.

\bibitem{Laffranchi2014}
M.~Laffranchi, L.~Chen, N.~Kashiri, J.~Lee, N.~Tsagarakis, and D.~G. Caldwell,
  ``{Development and control of a series elastic actuator equipped with a semi
  active friction damper for human friendly robots},'' \emph{Robotics and
  Autonomous Systems}, vol.~62, no.~12, pp. 1827--1836, 2014.

\bibitem{Ott2017}
M.~J. {Kim}, A.~{Werner}, F.~C. {Loeffl}, and C.~{Ott}, ``Enhancing joint
  torque control of series elastic actuators with physical damping,'' in
  \emph{IEEE International Conference on Robotics and Automation}, 2017, pp.
  1227--1234.

\bibitem{Focchi2016}
M.~Focchi, G.~A. Medrano-Cerda, T.~Boaventura, M.~Frigerio, C.~Semini,
  J.~Buchli, and D.~G. Caldwell, ``{Robot impedance control and passivity
  analysis with inner torque and velocity feedback loops},'' \emph{Control
  Theory and Technology}, vol.~14, no.~2, pp. 97--112, 2016.

\bibitem{Tognetti2005}
L.~J. Tognetti, ``{Improved design and performance of haptic two-port networks
  through force feedback and passive actuators},'' Ph.D. Dissertation, Georgia
  Institute of Technology, 2005.

\bibitem{Adams99}
R.~J. Adams and B.~Hannaford, ``Stable haptic interaction with virtual
  environments,'' \emph{IEEE Transactions on Robotics and Automation}, vol.~15,
  no.~3, pp. 465--474, 1999.

\bibitem{Haykin70}
S.~Haykin, \emph{Active network theory}.\hskip 1em plus 0.5em minus 0.4em\relax
  Reading, Mass. : Addison-Wesley Pub. Co, 1970.

\bibitem{Anderson89}
R.~J. {Anderson} and M.~W. {Spong}, ``Bilateral control of teleoperators with
  time delay,'' \emph{IEEE Transactions on Automatic Control}, vol.~34, no.~5,
  pp. 494--501, 1989.

\bibitem{Hannaford1989}
B.~Hannaford, ``{A design framework for teleoperators with kinesthetic
  feedback},'' \emph{IEEE Transactions on Robotics and Automation}, vol.~5,
  no.~4, pp. 426--434, 1989.

\bibitem{Fasse1987}
E.~D. Fasse, ``{Stability robustness of impedance controlled manupulator
  coupled to passive environments},'' M.Sc. Thesis, Massachusetts Institute of
  Technology, 1987.

\bibitem{Chen2009}
M.~Z. Chen and M.~C. Smith, ``A note on tests for positive-real functions,''
  \emph{IEEE Transactions on Automatic Control}, vol.~54, no.~2, pp. 390--393,
  2009.

\bibitem{Akritas2010}
A.~G. Akritas and P.~S. Vigklas, ``{Counting the number of real roots in an
  interval with Vincent's theorem},'' \emph{Bulletin Mathematique de la Societe
  des Sciences Mathematiques de Roumanie}, vol.~53, no.~3, pp. 201--211, 2010.

\bibitem{Hashtrudi-Zaad2001}
K.~Hashtrudi-Zaad and S.~E. Salcudean, ``{Analysis of Control Architectures for
  Teleoperation Systems with Impedance/Admittance Master and Slave
  Manipulators},'' \emph{The International Journal of Robotics Research},
  vol.~20, no.~6, pp. 419--445, 2001.

\bibitem{Colgatezwidth94}
J.~Colgate and J.~Brown, ``{Factors affecting the Z-Width of a haptic
  display},'' in \emph{IEEE International Conference on Robotics and
  Automation}, 1994, pp. 3205--3210.

\bibitem{Colgate95}
J.~Colgate, M.~Stanley, and J.~Brown, ``{Issues in the haptic display of tool
  use},'' in \emph{IEEE/RSJ International Conference on Intelligent Robots and
  Systems. Human Robot Interaction and Cooperative Robots}, vol.~3, 1995, pp.
  140--145.

\bibitem{Griffiths2006}
P.~G. Griffiths, R.~B. Gillespie, and J.~S. Freudenberg,
  ``{Performance/Stability Robustness Tradeoffs Induced by the Two-Port Virtual
  Coupler},'' in \emph{IEEE Symposium on Haptic Interfaces for Virtual
  Environment and Teleoperator Systems}, 2006, pp. 193--200.

\bibitem{Ogata2009}
K.~Ogata, \emph{Modern Control Engineering}, 5th~ed.\hskip 1em plus 0.5em minus
  0.4em\relax Upper Saddle River: Prentice Hall, 2009.

\end{thebibliography}
\normalsize

\appendix \label{sec:Appendix}

\begin{IEEEproof}[\hypertarget{pr:residue}{Proof} of Lemma~\ref{lem:residue}]
	If all the terms in any row of a Routh array are zero, then the characteristic equation has a pair of roots on the imaginary axis, and this special case may only occur at the odd-degree polynomial rows~\cite{Ogata2009}.
	\begin{equation*}
		\begin{array}{cccc}
			s^4 & a_4 & a_2 & a_0 \\
			s^3 & a_3 & a_1 \\
			s^2 & (a_2 a_3 - a_1 a_4) / a_3 & a_0 \\
			s^1 & \big(a_1(a_2 a_3 - a_1 a_4) - a_0 a_3^2\big)/(a_2 a_3 - a_1 a_4) & \\
			s^0 & a_0
		\end{array}
	\end{equation*}
	Since $a_i > 0$, the $s^3$-row cannot become zero.
	The only possibility is to have $a_1(a_2 a_3 - a_1 a_4) - a_0 a_3^2 = 0$ in the $s^1$-row, which completes the proof.
\end{IEEEproof}
\smallskip
\begin{IEEEproof}[\hypertarget{pr:residue-pos-real}{Proof} of Lemma~\ref{lem:residue-pos-real}]
	Since $ a_1(a_2 a_3 - a_1 a_4) = a_0 a_3^2 $, the impedance function $Z(s)$ has a pair of poles as given by Lemma~\ref{lem:residue}.
	Solving the auxiliary polynomial such that

	\begin{equation*}
		f_a(s) = \frac{a_2 a_3 - a_1 a_4}{a_3} s^2 + a_0 = 0,
	\end{equation*}

	\noindent the roots are found to be at $s = \pm j p$ where

	\begin{equation*}
		p = \sqrt{\frac{a_0 a_3}{a_2 a_3 - a_1 a_4}}.
	\end{equation*}

	For the residue, $r$, to be positive and real,
	\begin{IEEEeqnarray}{rCl}
		\text{Im}(r)&=0:\;& b_0 a_3^2 +b_4 a_1^2 - b_2 a_1 a_3 =\nonumber\\
		&&\hfill\frac{a_1 a_3^2 {(b_3 a_1 - b_1 a_3 )}}{a_2 a_3 -2 a_1 a_4 }\label{eq:Imr}\\
		\text{Re}(r)&>0:\;& \frac{b_1 a_3 - b_3 a_1 }{a_2 a_3 -2 a_1 a_4} > 0. \label{eq:Rer}
	\end{IEEEeqnarray}
	We notice that Eqn.~\eqref{eq:Rer} appears at the right-hand side of Eqn.~\eqref{eq:Imr}.
	Then, we can conclude the conditions given by Lemma~\ref{lem:residue-pos-real}.

	Similar analysis shows the same results for $s = -j p$.
\end{IEEEproof}
\smallskip
\begin{IEEEproof}[\hypertarget{pr:positive-thirdpoly}{Proof} of Lemma~\ref{lem:positive-thirdpoly}]
	Application of the Sturm's theorem results in the sign table below.

	\begin{table}[h!]
		\label{tab:sturmstable}
		\centering
		\begin{tabular}{c|c|c|c|c}
			\Xhline{1\arrayrulewidth}
			$ $       & $N_0$              & $N_1$              & $N_2$                    & $N_3$                   \\ \hline
			$x=0$        & $\text{sign}(p_0)$ & $\text{sign}(p_1)$ & $\text{sign}(\sigma_3 )$ & $\text{sign}(\sigma_1)$ \\
			$x\to\infty$ & $\text{sign}(p_3)$ & $\text{sign}(p_3)$ & $\text{sign}(\sigma_2 )$ & $\text{sign}(\sigma_1)$ \\
			\Xhline{1\arrayrulewidth}
		\end{tabular}
	\end{table}

	\noindent where
	\begin{IEEEeqnarray}{rCl}
		\sigma_1&=&-4 p_1 \sigma_2^2 - 3 p_3 \sigma_3^2 + 4 p_2 \sigma_3 \sigma_2 \nonumber\\
		\sigma_2&=&p_2^2-3p_3p_1 \nonumber\\
		\sigma_3&=&p_1p_2-9p_0p_3. \nonumber
	\end{IEEEeqnarray}

	Non-negativeness of the polynomial $p(x)$ allows roots of even multiplicity on the x-axis.
	However, proving positiveness of the polynomial provides the non-negativeness at the limits of the derived conditions.
	Therefore, without loss of generality, the following proof ensures $p(x)$ does not have real roots.

	Non-negativeness of $ p(x) $ at the boundaries of $ x\in[0, \infty) $ requires that $ p_0\geq0 $.
	Given $ p_3 > 0 $, all possible conditions that will result in an equal number of sign changes in the Sturm's sequence may be summarized as follows.

	\begin{enumerate}
		[label=($\ast$)]
		\item[(1)] if $ p_0\geq0 $ and $\sigma_2 \leq 0$ and $(\sigma_3<0 \lor \sigma_1<0)$,
		\item[(2)] if $ p_0\geq0 $ and $\sigma_2 > 0$ and one of the following holds
		\begin{enumerate}
			[label=($\ast$)]
			\item[(i)]  $p_1>0 \land \sigma_3 > 0$,
			\item[(ii)] $(p_1>0 \lor \sigma_3<0) \land \sigma_1 < 0$.
		\end{enumerate}
	\end{enumerate}

	In Condition (1), rearranging $\sigma_2 \leq 0$ as $0 \leq p_2^2 \leq 3p_1 p_3$ implies $p_1 \geq 0$ and $-\sqrt{3p_1 p_3 } \leq p_2 \leq \sqrt{3p_1 p_3 }$.
	To simplify the analysis, we can consider positive and negative cases of $p_2$ separately.
	If $p_2 \leq 0$ then $\sigma_3 = p_1 p_2 - 9p_0 p_3 < 0$, which is sufficient for Condition (1) to hold.
	On the other hand, if $p_2 > 0$ then we can rewrite $\sigma_1 < 0$ as $\sigma_3 > (4 p_1 \sigma_2^2 + 3 p_3 \sigma_3^2)/(4p_2\sigma_2)$.
	Note that, the right hand side of the inequality is always negative since $p_1 \geq 0$, $p_2 > 0$ and $\sigma_2 \leq 0$.
	Therefore, Condition (1) and positive realness are satisfied regardless of the signs of $\sigma_1$ and $\sigma_3$ if

	\begin{equation}
		\label{eq:3rd-poly-1}
		-\sqrt{3p_1 p_3 } \leq p_2 \leq \sqrt{3p_1 p_3 }.
	\end{equation}

	In Condition (2), rearranging $\sigma_2 > 0$ as $p_2^2 > 3p_1 p_3$ implies  $p_2 < -\sqrt{3p_1 p_3 }$ or $p_2 >\sqrt{3p_1 p_3 }$ if $p_1 > 0$; otherwise, $p_2 \in \real$.

	Condition (2-i) requires $p_2 > 9 p_0 p_3 / p_1$.
	However, Condition (2) and Eqn.~\eqref{eq:3rd-poly-1} may be merged as follows

	\begin{equation*}
		p_1 \geq 0 \land -\sqrt{3p_1 p_3 } \leq p_2,
	\end{equation*}

	\noindent which is sufficient to satisfy the requirement in Condition (2-i).
	Condition (2-ii) is equal to $\sigma_3 < 0 \land \sigma_1 < 0$, since $p_1 > 0$ does not introduce any additional restriction.
	This completes the proof.
\end{IEEEproof}
\smallskip
\begin{IEEEproof}[\hypertarget{pr:thm-5}{Proof} of Theorem~\ref{thm:SDEA-2port-pass}]
	Condition (\emph{a}) of Theorem~\ref{thm:2port-pass} requires the Routh-Hurwitz test on the diagonal elements of \emph{h}-matrix.
	$h_{22}$ is selected as passive; therefore, it is stable.
	The characteristic equation of $h_{11}$ is of the form considered in Lemma~\ref{lem:RH}.
	Then, the system is stable if and only if the following condition holds.

	\small
	\begin{IEEEeqnarray}{r}
		\label{eq:routh-cond}
		\frac{K_f \mu \nu}{B_f \mu \nu + K_f (\mu + \nu)}((B + P_m) + B_f(\alpha +P_m P_f ))^2 \leq \hfill\\
		B_f (\alpha + P_m P_f)[I_m + (\alpha +P_m P_f) K_f + B_f P_m P_f(\mu +\nu)] \nonumber\\
		+ \bigg[\kappa_3 +\frac{B_f P_m P_f}{K_f} \bigg((B + P_m)(\mu + \nu) - M \mu \nu\bigg)\bigg] K_f \nonumber\\
		+ I_m (B + P_m). \nonumber
	\end{IEEEeqnarray}
	\normalsize

	In the case of Eqn.~\eqref{eq:routh-cond} is satisfied as equality, $h_{11}$ has a pair of conjugate poles at the imaginary axis.
	For Condition~(\emph{b}), the following conditions ensure positive and real residues at those poles.
	\begin{IEEEeqnarray}{rcl}
		\label{eq:residue-cond}
		0 &< &\beta = (K_f \rho +B_f I_m)(\rho + B_f \alpha + B_f P_m P_f) \nonumber\\
		&&- B_f M(B_f I_m I_f + K_f P_m P_f (\mu +\nu)) \IEEEyesnumber\IEEEyessubnumber
	\end{IEEEeqnarray}
	\begin{IEEEeqnarray}{rcl}
		\big(K_f I_m a_3 &-& a_1 (B_f \rho + K_f M)\big)a_3^2 = \IEEEyessubnumber\\
		&\beta \big(&a_3 (I_m + K_f \alpha + P_m P_f (K_f +B_f (\mu +\nu)))\nonumber \\
		&&-M (2 B_f I_m I_f +2 K_f P_m P_f (\mu +\nu))\big). \nonumber
	\end{IEEEeqnarray}
	For Condition (\emph{c-i}), positive realness of $h_{22}$ is already assured by selection.
	On the other hand, $\textrm{Re}(h_{11}) $ is reduced to the inequality below by Lemma~\ref{lem:ReHs}, followed by the substitution of $\omega^2$ by $x$.
	\begin{IEEEeqnarray}{rcl}
		\label{eq:Reh11}
		0\; &\leq &\; B_f M^2 x^4 + B_f \left(B_f \kappa_3 +(B + P_m )^2 - 2 I_m M \right) x^3\nonumber\\
		&&+ \left(K_f^2 \kappa_3 +B_f I_m^2 + B_f^2 I_m \kappa_1 \right)x^2 + I_m K_f^2 \kappa_1 x,
	\end{IEEEeqnarray}
	\noindent where
	\begin{IEEEeqnarray}{rcl}
		\kappa_1 &=& P_f I_m - B I_f \IEEEyessubnumber\\
		\kappa_2 &=& B + P_m - M (\mu + \nu) \IEEEyessubnumber\\
		\kappa_3 &=& \alpha (B + P_m) + P_m P_f \kappa_2. \IEEEyessubnumber
	\end{IEEEeqnarray}

	Eqn.~\eqref{eq:Reh11} is of the form $r(x) = x\,(r_3 x^3 + r_2 x^2 + r_1 x + r_0)$ for $x \geq 0$.
	Then, Lemma~\ref{lem:positive-thirdpoly} ensures $r(x) \geq 0$ for $x \geq 0$ providing the necessary and sufficient conditions for $r_3 x^3 + r_2 x^2 + r_1 x + r_0$.
	Lemma~\ref{lem:positive-thirdpoly} requires that $r_0 \geq 0$, which implies

	\begin{IEEEeqnarray}{c}
		\label{eq:SDEA-2port-proof-r0}
		I_f \leq \frac{I_mP_f}{B}. \IEEEyesnumber
	\end{IEEEeqnarray}

	Immediately following Lemma~\ref{lem:positive-thirdpoly}, we find the inequalities given in \hyperlink{thm:SDEA-2port-pass-c-i}{Condition} \emph{(c-i)} of Theorem~\ref{thm:SDEA-2port-pass}.

	Following the same steps as presented above, Condition~\emph{(c-ii)} leads to the polynomial below.
	\begin{IEEEeqnarray}{rCl}
		\label{eq:poly-two-port-case-cii}
		0 & \leq & \tau_1 b_{22} M^2 x^5 + (4 b_{22} r_2 +b_{22}^2 \tau_2 - k_{22}^2 M^2) x^4 \nonumber\\
		&& + \left(4 b_{22} r_1 + k_{22}^2 \tau_2 - b_{22}^2 (I_m + \alpha K_f)^2 \right) x^3 \\
		&& + \left(4 b_{22} K_f^2 I_m \kappa_1 - k_{22}^2 (I_m +\alpha K_f)^2 \right) x^2, \nonumber
	\end{IEEEeqnarray}
	\noindent where
	\begin{IEEEeqnarray}{rCl}  \label{eq:bfrequirement_gen}
		\tau_1 &=& 4 B_f - b_{22} \IEEEyessubnumber\\
		\tau_2 &=& 2 M \left(I_m + \alpha K_f \right) -\left(B + P_m + \alpha B_f \right)^2. \IEEEyessubnumber
	\end{IEEEeqnarray}

	Eqn.~\eqref{eq:poly-two-port-case-cii} is of the form $t(x) = x^2\,(t_3 x^3 + t_2 x^2 + t_1 x + t_0)$ for $x \geq 0$.
	Since Lemma~\ref{lem:positive-thirdpoly} assumes $t_3 > 0$ and requires $t_0 \geq 0$ we have
	\begin{IEEEeqnarray}{rCl}
		\label{eq:alphaval}
		0 & < & b_{22} \leq 4 B_f \IEEEyesnumber\IEEEyessubnumber\label{eq:b22requirement}\\
		0 & \leq & t_0 = 4 b_{22} K_f^2 I_m \kappa_1 - k_{22}^2 (I_m + \alpha K_f)^2. \IEEEyessubnumber\label{eq:SDEA-2port-proof-t0}
	\end{IEEEeqnarray}

	Although the condition $t_3 > 0$ allows negative $b_{22}$ values, $t_0 \geq 0$ eliminates the non-positive region.
	Note that, in Eqn.~\eqref{eq:SDEA-2port-proof-t0}, the first monomial should compensate for the negative effect of the second.
	Then, $\kappa_1$ must be greater than some positive constant, dictating more strict condition than Eqn.~\eqref{eq:SDEA-2port-proof-r0}.

	Immediately following the other conditions of Lemma~\ref{lem:positive-thirdpoly}, we find the inequalities given in \hyperlink{thm:SDEA-2port-pass-c-ii}{Condition} \emph{(c-ii)} of Theorem~\ref{thm:SDEA-2port-pass}. This completes the proof.
\end{IEEEproof}

\end{document}